\documentclass{article}

\usepackage[final]{neurips_2024}

\usepackage[utf8]{inputenc} % allow utf-8 input
\usepackage[T1]{fontenc}    % use 8-bit T1 fonts
\usepackage{hyperref}       % hyperlinks
\usepackage{url}            % simple URL typesetting
\usepackage{booktabs}       % professional-quality tables
\usepackage{amsfonts}       % blackboard math symbols
\usepackage{nicefrac}       % compact symbols for 1/2, etc.
\usepackage{microtype}      % microtypography
\usepackage{xcolor}         % colors

\usepackage{graphicx} 
\usepackage{amsmath}
\usepackage{booktabs}
\usepackage{subfigure}

\newcommand{\tool}{CE$^2$}
\newcommand{\toolegc}{CE$^2$-G}
\newcommand{\toolwithoutpeg}{CE$^2$-noPEG}
\usepackage{wrapfig}
\usepackage[toc,page,title]{appendix}
\usepackage{minitoc}
\usepackage[noend]{algpseudocode}
\usepackage{algorithm}

\title{Exploring the Edges of Latent State Clusters for Goal-Conditioned Reinforcement Learning}

\author{%
  Yuanlin Duan\\
  Rutgers University\\
  \texttt{yuanlin.duan@rutgers.edu} \\
  \And
  Guofeng Cui \\
  Rutgers University \\
  \texttt{gc669@cs.rutgers.edu} \\
  \And
  He Zhu \\
  Rutgers University \\
  \texttt{hz375@cs.rutgers.edu} \\
}

\begin{document}
\include{pythonlisting}

\maketitle

\begin{abstract}
  Exploring unknown environments efficiently is a fundamental challenge in unsupervised goal-conditioned reinforcement learning. While selecting exploratory goals at the frontier of previously explored states is an effective strategy, the policy during training may still have limited capability of reaching rare goals on the frontier, resulting in reduced exploratory behavior.
  We propose "Cluster Edge Exploration" (\tool{}), a new goal-directed exploration algorithm that when choosing goals in sparsely explored areas of the state space gives priority to goal states that remain accessible to the agent. The key idea is clustering to group states that are easily reachable from one another by the current policy under training in a latent space and traversing to states holding significant exploration potential on the boundary of these clusters before doing exploratory behavior. In challenging robotics environments including navigating a maze with a multi-legged ant robot, manipulating objects with a robot arm on a cluttered tabletop, and rotating objects in the palm of an anthropomorphic robotic hand, \tool{} demonstrates superior efficiency in exploration compared to baseline methods and ablations.
\end{abstract}

\section{Introduction}
In recent years, Goal-Conditioned Reinforcement Learning (GCRL)~(\cite{andrychowicz2017hindsight}) has emerged as a powerful paradigm for training agents to accomplish diverse tasks in complex and dynamic environments. GCRL enables agents to learn goal-directed behaviors, allowing them to achieve specific objectives in a flexible and adaptive manner. However, a central challenge in GCRL lies in guiding agents to effectively explore their environment during training.
%particularly when the goals they need to achieve are revealed only at test time. 
The exploration problem in GCRL can be viewed as the task of setting goals for the agent during training to guide the agent's environment navigation to collect exploratory data that improves its learning process. In this paper, we address this critical challenge by proposing a novel strategy for selecting exploration-inducing goals in GCRL.
%goal-directed exploration in GCRL %at training time to facilitate effective exploration and learning.
%select exploration-inducing goals in GCRL?

Because goal-conditioned policies excel at reaching states encountered frequently during training, a simple strategy is setting goals in less-visited areas of the state space to broaden the range of reachable states. However, throughout training, goal-conditioned policies may encounter difficulties in reaching arbitrary goals. For example, when instructed to navigate to an unexplored section of a maze environment, a novice agent might instead revisit a previously traversed area that provides low exploration value. To address this shortcoming, the environment exploration procedure must set up additional mechanisms to filter out unreachable goals. A common strategy in the literature is to select goals at the frontier of previously explored states and launch an exploration phase immediately after these goals are achieved,
%followed by the initiation of an exploration phase once these goals are attained, 
adhering to a Go-Explore principle~(\cite{ecoffet2019go}). For example, Skewfit~(\cite{pong2019skew}) estimates state densities and selects goals at the frontier from the replay buffer in inverse proportion to their density. Similarly, MEGA~(\cite{pitis2020maximum}) uses kernel density estimates (KDE) of state densities and selects frontier goals with low density from the replay buffer. However, precisely identifying the frontier of known states can be challenging with these heuristics. Even once the frontier is identified, the policy during training may still have limited capability of reaching rare goals on the frontier, resulting in reduced exploratory behavior.

To address the aforementioned challenge, we propose a new goal-directed exploration algorithm, \tool{} (short for "Cluster Edge Exploration"). When choosing goals in sparsely explored areas of the state space, \tool{} gives priority to goal states that remain accessible to the agent. For this purpose, our key idea is clustering to group known states that are easily reachable from one another by the current policy under training, and traversing to states holding significant exploration potential on the boundary of these clusters before doing exploratory behavior. In this way, our method accounts for the capability of the current policy for exploratory goals.
%reaching states located on the, and then initiating exploration from these states. 
First, a state cluster likely represents part of the state space where the training policy is familiar with. Second, given the easy accessibility of states within each cluster by the training policy, the agent's capability extends to reaching states even at cluster boundaries. Moreover, less explored regions naturally reside adjacent to the periphery of state clusters. This Go-Explore strategy enables the agent to progressively broaden the coverage of each state cluster to effectively explore a novel environment.
We instantiate \tool{} in the context of model-based GCRL, demonstrating how learned world models can facilitate clustering environment states that are easily reachable from one another by the training policy in a latent space. We validate the effectiveness of \tool{} in challenging robotics scenarios, including navigating a maze with a multi-legged ant robot, manipulating objects with a robot arm on a cluttered tabletop, and rotating objects in the palm of an anthropomorphic robotic hand. In each scenario, \tool{} exploration results in more efficient training of adaptable GCRL policies compared to baseline methods and ablations. 

\section{Problem Setup and Background}
\label{sec:psb}

Our work focuses on the exploration problem in unsupervised goal-conditioned reinforcement learning (GCRL) settings. In this section, we set up notation and preliminary concepts.

\textbf{GCRL}. A goal-conditioned Markov decision process (MDP) is defined by the tuple ($S$, $A$, $G$, $T$, $\eta$) where the state space $S$ defines the set of all possible agent's observations into the environment, the action space $A$ defines all possible actions that the agent can take in each state, $G$ is the set of all possible goals that the agent may aim to achieve in the environment, and the transition function $T$ describes the probability of transitioning from one state to another given an action. It is defined as $T(s' | s, a)$, where $s' \in S$ is the next state, $s \in S$ is the current state, and $a \in A$ is the action taken. $\eta : S \rightarrow G$ is a tractable mapping function that maps a state to a specific goal. A goal-conditioned $\pi(a | s, g)$ represents the agent's strategy for selecting actions based on states and goal commands, indicating the probability of taking action $a$ in state $s$ given goal command $g \in G$. 
In this paper, for ease of presentation, we assume $S = G$ and $\eta$ is an identify function.

\begin{wrapfigure}[13]{r}{0.5\textwidth}
\vspace{-0.2cm}
  \centering
  \includegraphics[width=0.5\textwidth]{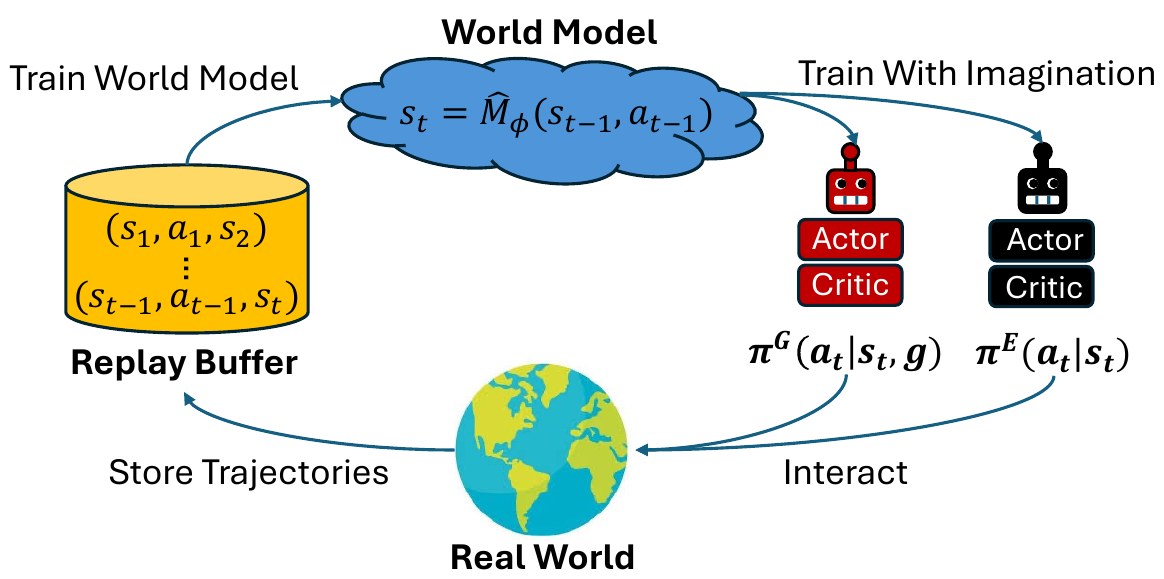}
  \caption{Model-based GCRL Framework}
  \label{fig:mbrl}
\end{wrapfigure}

%the goals they need to achieve are revealed only at test time
Our goal is to develop agents capable of unsupervised exploration when dropped into an unknown environment. During the unsupervised exploration stage, there are no predefined tasks or goals. The agent sets its own goal command $g \in G$ as it explores the environment. Following this exploration phase, a successful agent should be able to navigate to a wide range of previously unknown goal states in the environment upon goal commands. %
%
%The reward received by the agent $R(s, a, g)$ is 1 only if it achieves $g$ at $s$ and is 0 otherwise.

\textbf{Model-based GCRL}. Model-based reinforcement learning (MBRL) is an approach where an agent learns a model of the environment's dynamics to predict future states, enabling more efficient policy learning. Fig.~\ref{fig:mbrl} shows the general MBRL framework. We use the world model structure $\hat{M}$ of Dreamer (\cite{hafner2019dream, hafner2019learning, hafner2020mastering, hafner2023mastering}) to learn real environment dynamics as a recurrent state-space model (RSSM). We provide a detailed explanation of the network architecture and working principles of the RSSM in Appendix~\ref{subs: rssm}. Particularly, we consider \textbf{GC-Dreamer} (goal-conditioned Dreamer) as a baseline. In GC-Dreamer, the goal-conditioned agent $\pi^G(a | s, g)$ samples goal commands $g \in G$ from a given environment goal distribution $p_g$ to collect trajectories in the real world. These trajectories are used to train the world model $\hat{M}$, and subsequently, $\pi^G$ is trained on imagined rollouts generated by $\hat{M}$, with these two steps run in alternation. The reward function used to train $\pi^G$ is determined by a temporal distance network $D_t$ (see below).

\textbf{Go-Explore}. In unsupervised GCRL, the goal distribution $p_g$ is only revealed at test time. "Go-Explore"~(\cite{ecoffet2019go,pislar2021should,tuyls2022multi,hu2023planning}) is a popular mechanism tailored for long-term GCRL scenarios that require extensive exploration. The Go-Explore methodology splits each training episode into two distinct phases: the "Go-phase" and the "Explore-phase". In the "Go-phase", the agent is guided to an "interesting" goal $g$~(\cite{pong2019skew,pitis2020maximum}) (e.g., states rarely encountered in the replay buffer) by the GCRL policy $\pi^G$, reaching a final state $s_{T}$. Subsequently, the "Explore-phase" kicks in, with an undirected exploration policy $\pi^E$ taking over from $s_{T}$ for the remaining timesteps. This exploration policy is optimized to maximize an intrinsic exploration reward~(\cite{bellemare2016unifying,pathak2017curiosity,burda2018exploration,sekar2020planning}) (e.g., to explore less familiar areas of the environment that the world models have not adequately learned). %This dual-phase structure of training episodes has been demonstrated to enhance exploration~\cite{pislarSOBS22}.

%And we provide a detailed explanation of the network architecture and working principles of the RSSM in Appendix~\ref{subs: rssm}, along with the definition of mathematical symbols for some components that will be used later.
%In MBRL, the world model $\hat{M}$ is trained using trajectories sampled from the real world by agent, 
%and then use the imagined rollouts of trained $\hat{M}$ to learn a better policy $\pi^G$ and $\pi^E$.

Recently, Go-Explore has been integrated with model-based unsupervised GCRL~(\cite{mendonca2021discovering,hu2023planning}), as depicted in Fig.~\ref{fig:mbrl}. In addition to the goal-conditioned policy $\pi^G(a | s, g)$, an exploration policy $\pi^E(s)$ is introduced into the model-based GCRL framework. The agent's training process involves learning the following components:
%the a goal-conditioned MBRL framework, called Latent Explorer Achiever (LEXA) which consists of a explorer and goal reaching policy, denoted as $\pi^E$ and $\pi^G$ respectively.
\begin{align*}\label{eq:worldmodel}
  & \text{World Model:} & & \hat{M}(s_t | s_{t-1}, a_{t-1}) & & & & \\
  & \text{Exploration policy:} & & \pi^E(s_t) & & \text{Goal Reaching policy:} & & \pi^G(s_t, g) \\
  & \text{Exploration value:} & & V^E(s_t) & & \text{Goal Reaching value:} & & V^G(s_t, g) 
\end{align*}
where both $\pi^G$ and $\pi^E$ are trained using the model-based actor-critic algorithm in Dreamer~(\cite{hafner2020mastering}). They are entirely trained with the imagined rollouts of the world model $\hat{M}$ to maximize the accumulated rewards $\sum_t r^G_t$ and $\sum_t r^E_t$, respectively.
%It is worthy to note that the $\pi^G$ is goal conditioned, while the $\pi^E$ is undirected exploration policy.
The explorer reward $r^E$ encourages exploration by leveraging the Plan2Explore~(\cite{sekar2020planning}) disagreement objective, which motivates the agent to seek states that induce discrepancies among an ensemble of world models.
In contrast, the goal-reaching reward $r^G$ is driven by the self-supervised temporal distance objective $D_t$ (\cite{mendonca2021discovering}), 
which reinforces the policy to minimize the action steps required to transition from the current state $s$ to a sampled goal state $g$ in an imagined rollout, i.e., $r^G(s, g) = -D_t(\Psi(s), \Psi(g))$. The temporal distance network $D_t$ predicts the anticipated number of action steps needed to transition from $s$ to $g$. It is trained by extracting pairs of states $s_t$ and $s_{t+k}$ from an imagined rollout generated by $\hat{M}$ and predicting the distance $k$ as shown in Equation~\ref{eq: temporal_distance_reward} where $H$ is the total length of the imagined rollout:
\begin{equation}\label{eq: temporal_distance_reward}
  D_t\big(\Psi(s_{t}), \Psi(s_{t+k})\big) \approx k/H
\end{equation}
Here, $\Psi$ is a learned function for state embeddings in the world model (we assume $S = G$ in the paper). Further details on the training procedure of $D_t$ can be found in Appendix~\ref{subs: Dt-training}.

\tool{} aims to address the core challenge in the Go-Explore mechanism: how do we select an interesting goal command $g$ at the frontier of known states with high exploration potential and effectively guide the agent to $g$?

\section{State Cluster Edge Exploration}

% Upon establishing the background and prerequisites, the objective of this section is to introduce our approach, 
% a novel and robust reinforcement learning exploration algorithm. In sparse reward environments, 
% devising an exploration strategy that balances breadth of exploration while achieving satisfactory levels of exploration speed 
% and depth is challenging. Currently, although there exist some exploration strategies that perform reasonably well 
% in certain sparse reward environments, they either rely on premises such as environments that can be reset to specific states 
% or are based on particular environment characteristics, such as the narrow exploration space. 
% They lack general applicability and cannot achieve good performance in the majority of sparse reward environments, 
% including challenging exploration environments like the 3-block stacking task. 

%goal-directed policy efficiently traverses to the frontier of known states before doing exploratory behavior

%Our algorithm is grounded in an innovative idea that argues against pure randomness or simplistic planning strategies as the basis for the exploration process.

The major limitation in existing Go-Explore approaches, such as those described in ~(\cite{pong2019skew, pitis2020maximum}) is that the policy under training can struggle to reach heuristically chosen rare goals at the frontier of known states~(\cite{hu2023planning}). This difficulty arises because the goal commands are selected without a systematic method to filter out unachievable goals for the agent, leading to diminished exploratory behavior. In \tool{}, when choosing goals in sparsely explored areas of the state space in the "Go-phase", our method gives priority to goal states that remain accessible. For this purpose, the key idea is clustering to group states that are easily reachable from one another by the current policy under training in a latent space, and selecting states holding significant exploration potential on the boundary of these clusters as the "interesting" goals to explore. In Sec.~\ref{sec:latentspace}, we discuss how to learn a latent space that can represent the reachability relationships between environment states. In Sec.~\ref{sec:clustering}, we explain how this latent space can be used to cluster states in the replay buffer that are easily reachable from one another. In Sec.~\ref{sec:selecting}, we demonstrate how the agent can be brought to interesting states on the boundary of latent state clusters to effectively explore its environment. 

\subsection{Latent Space Learning}
\label{sec:latentspace}

%In Model Based Reinforcement Learning, the agent learns a world model $\hat{M}$ using trajectories sampled from the real world. 
%This world model is capable of predicting the next state of the environment after taking a specific action in a given state. 
Typically, during the learning process of a world model $\hat{M}$ as a neural network, an essential step involves encoding states from the original observation space into a latent space using an encoder, which can then be decoded back to the original observation space by a decoder. This latent space is subsequently used to learn the dynamic model of the real environment~\cite{hafner2019dream, hafner2020mastering}.
%The optimization of this latent space is aimed at ensuring that when a state is encoded into the latent space, 
%it can be maximally decoded back to the original observation space by the decoder. 
%This property of the latent space is crucial and fundamental as it enables the agent to comprehend the structure of the observation space to some extent, 
%facilitating more efficient learning. However, 

In \tool{}, we additionally require the latent space can express the temporal distance between different states. In other words, we aim for the distances between various states in the latent space to represent the number of steps required to transition from one another in the real environment (after decoding) by the training policy. Therefore, the loss function of training the latent space in \tool{} comprises two components. 
The first component is the reconstruction loss $\mathcal{L}_{rec}$, akin to the latent space loss function in Dreamer framework~(\cite{hafner2019dream, hafner2020mastering}). 
It captures the association between the latent space and the re-decoding to the observation space, 
%along with predicting the reward function, discount factor, 
along with predicting dynamic transition in the latent space. 
We introduce a second loss term $\mathcal{L}_{dt}$ that leverages the temporal distance network $D_t$ in Equation~\ref{eq: temporal_distance_reward} to guide the learning of the latent space structure.
For any pair of states $(s1, s2)$ sampled from the replay buffer, the $\mathcal{L}_{dt}$ loss function is formulated as follows ($\Psi$ is a learned function for state embeddings in the world model):
\begin{equation}
\mathcal{L}_{dt} = (\| \Psi(s_1) - \Psi(s_2) \|_2^2 - \frac{1}{2}(D_t(\Psi(s_1),\Psi(s_2)) + D_t(\Psi(s_2), \Psi(s_1))))^2
\label{eq:dt_loss}
\end{equation}
\begin{equation}
\mathcal{L}_{latent} = \mathcal{L}_{rec} + \mathcal{L}_{dt}
\end{equation}
%In Equation \ref{eq:dt_loss}, $f_E$ represents the encoder of $\hat{M}$, $g_1$ and $g_2$ are any two states from replay buffer, and $D_t$ is a temporal distance estimation network.
%
We use the loss function $\mathcal{L}_{latent}$ to supervise the training of the latent space. The trained latent space provides the agent with a deeper understanding of the real environment, 
where states that are easily reachable from one another in the real environment are closer in proximity within the latent space. 
% By incorporating this constraint and jointly optimizing it, the mapping between observation space and latent space is no longer arbitrary, 
% thus imbuing the latent space with a more meaningful structure. Interpolations and samplings between encodings in the latent space yield meaningful results. 
% This facilitates our next steps, including clustering within the latent space and sampling encodings at the edges of clusters that hold potential exploration value. 
% Additionally, we have the freedom to transform encoding sampled back to the observation space for evaluation and planning purposes.

\subsection{Latent State Clustering}
\label{sec:clustering}

To identify the frontier of known states, \tool{} conducts state clustering to group states in the replay buffer. States that are easily reachable from one another are classified in the same cluster in the latent space by Gaussian Mixture Models (GMMs), based on the temporal distances between the encoded states. GMMs are probabilistic models that assume all data points are generated by a mixture of a finite number of Gaussian Distributions.
%The number of clusters that the latent space can express varies for different environments, depending on the spatial characteristics and task complexity. 
%Extracting the desired clusters from a high-dimensional and extensive latent space is an important challenge.
We initialize the Gaussian models in the latent space with $N_{c}$ trainable latent centroids $c = \{c_1, \dots, c_{N_{c}}\}$ and a shared variance $\sigma$, where $N_{c}$ represents the desired number of clusters. These $N_{c}$ latent centroids are initialized by applying the Farthest Point Sampling (FPS) algorithm(\cite{eldar1997farthest}) to select a representative subset of states from a batch of data sampled from the replay buffer. We provide a detailed description of the FPS algorithm in Appendix~\ref{subs: FPS}. After initialization, we optimize the clustering model by maximizing the Evidence Lower Bound (ELBO) iteratively on sampled batches from the replay buffer with a uniform prior $p(c)$ to scatter out the latent centroids~(\cite{zhang2021world}): 
\begin{equation}
  \log{p(z = \Psi(s))} \geq \mathbb{E}_{q(c|\Psi(s))}[\log{p(\Psi(s)|c)}] - D_{KL}(q(c|\Psi(s))||p(c)) 
\end{equation}
where $p$ and $q$ are represented as Gaussian distributions within the GMMs. $q(c|\Psi(s))$ is the postior distribution over $c$ (the clusters) given an encoded state $\Psi(s)$. 
$\log{p(\Psi(s)|c)}$ is the distribution donating the probability of the encoded state $\Psi(s)$ in cluster $c$. $p(c)$ is the prior distribution of the weight of clusters in GMMs. 
For each round of optimization, we increase the probability of the sampled batches in GMMs by updating the weight of each cluster $c$ in GMMs and the mean and variance of them.
%We aim for the learned clusters to express key state regions within the latent space relevant to the task, as well as the boundaries of these crucial state regions. We initialize Gaussian Mixture Models (GMMs) using batch trajectories stored in the Replay Buffer. 
%If these batches originate from exploration trajectories, the agent is encouraged to prioritize exploration breadth, thereby swiftly navigating unknown areas within the environment. %Conversely, if these batch data stem from the agent's goal-conditional policy trajectories targeting task final goals $G$, the agent is prompted to focus more on exploring the edges of crucial state regions along the path to the $G$, thus rapidly enhancing the agent's capability to achieve the ultimate environmental goal. In this manner, our Gaussian Mixture Models can concurrently express the agent's demands for both breadth and depth of exploration within the environment.
%the agent is prompted to focus more on exploration initiating from the edges of latent state clusters along the path to the states in $p_g$. The intuition is that ensures exploration novelty while gradually increasing the depth of exploration around the space surrounding the path to the final goal $G$.

\subsection{Exploring the Boundaries of Latent State Clusters}
\label{sec:selecting}

Assuming we have already trained $N_c$ state clusters in the latent space, each representing part of the state space where the goal-conditioned policy under training is familiar with, how can we utilize these state clusters to plan an exploration strategy? 
\tool{} selects goal states at the edges of these latent state clusters for exploration because (1) less explored regions are naturally adjacent to these boundaries, and (2) given the easy accessibility between states within each cluster by the training policy, the agent's capability extends to reaching states even at the cluster boundaries.

%we demonstrate how a learned world model can be used to select goals states with the highest exploration potential from the latent state cluster edges.

We outline our exploration algorithm in Algorithm~\ref{alg:CE2}. At line~\ref{alg:CE2:sample}, it samples $N_{candidate}$ latent states as $S_{candidate}$ from GMMs. A higher sampling quantity ensures sampling from more states at the edges of the clusters. We set $N_{candidate} = 1000$ in \tool{}. 
We compute the total probability of each latent state $\hat{s} \in S_{candidate}$ in the Gaussian mixture model, given by the formula:
\begin{equation}
  p(\hat{s}) = \sum_{i=1}^{N_{c}} \beta_i \mathcal{N}(\hat{s} | c_i, \sigma)
\label{eq: GMM}
\end{equation}
In this formula, \(\beta_i\) are the mixture weights satisfying \(\beta_i \geq 0\) and \(\sum_{i=1}^{N_{c}} \beta_i = 1\), \(\mathcal{N}(\hat{s} | c_i, \sigma)\) represents the \(i\)-th Gaussian distribution with mean \(c_i\) and the shared standard deviation \(\sigma\). 
At line~\ref{alg:CE2:edge}, we select $N_{edge}$ latent states with the lowest total probability from $S_{candidate}$ by Equation~\ref{eq: GMM} as a set $S_{edge}$. Intuitively, these states reside on the edges of the latent state clusters and, therefore, induce a set of a goal commands $G_{edge} = \{ \eta(f_D(\hat{s})) | \hat{s} \in S_{edge}\}$ that may be used for the "Go-phase" for Go-Explore, where $f_D$ is the state decoder and $\eta$ is the goal mapping function.
\begin{wrapfigure}[9]{r}{0.57\textwidth}
\vspace{-0.5cm}
\scalebox{0.8}{
\begin{minipage}{0.7\textwidth}
\begin{algorithm}[H]
\caption{Cluster Edge Exploration(\tool{})}
\label{alg:CE2}
\begin{algorithmic}[1]
\State $D_{exp} \leftarrow\{\}$
\For{episode $i = 1$ to $N_{\tau}$}
    \State $S_{candidate} \leftarrow$ Sample $N_{candidate}$ points from $GMM$
    \label{alg:CE2:sample}
    \State $G_{edge} \leftarrow$ $N_{edge}$ states in $S_{candidate}$ with the smallest total probability based on Equation~\ref{eq: GMM}. \label{alg:CE2:edge}
    \State $g^E \leftarrow argmax_{g\in G_{edge}}\ P^E(g)$ through imagination with $\hat{M}$ \label{alg:CE2:ge}
    \State $\tau \leftarrow \textit{GO-EXPLORE}(g^E, \pi^G, \pi^E)$
    \State $D_{exp} \leftarrow D_{exp} \cup \tau$
    \EndFor
\end{algorithmic}
\end{algorithm}
\end{minipage}
}
\end{wrapfigure}
However, randomly picking a goal command from $G_{edge}$ overlooks whether the policy can exactly navigate the agent to the sampled goal in the real environment. Although determining the exact outcome of the policy without execution is impractical, similar to PEG (\cite{hu2023planning}), we can leverage the world model to provide an approximation of the exploration potential $P^E(g)$ of a goal command $g$:
% Next, to evaluate the exploration potential $P^E$ of each goal command from $G_{edge}$, we use the method from the PEG(\cite{hu2023planning}), which leverages the world model to approximate the exploration potential.
\begin{equation}
  \hat{p}_{\pi^G(\cdot \vert \cdot,g)(\tau)} = p(s_0)[\prod^{T}_{t=1} \hat{M}(s_t|s_{t-1},a_{t-1})\pi^G(a_{t-1}|s_{t-1},g)]
\label{eq:exp_potential_3}
\end{equation}
\begin{equation}
  \begin{aligned}
  \hspace{1.7em} P^E(g) &= \mathbb{E}_{p_{\pi^G(\cdot \vert \cdot,g)(s_T)}}[V^E(s_T)]
  &\approx \frac{1}{K}\sum_{k}^{K}V^E(s_T^k)  \text{\qquad where } s^k_T \sim \hat{p}_{\pi^G(\cdot \vert \cdot,g)(\tau)}\\
  \end{aligned}
\label{eq:exp_potential_2}
\end{equation}
%
%
% \begin{equation}
%   P^E(g) = \mathbb{E}_{p_{\pi^G(\cdot \vert \cdot,g)(s_T)}}[V^E(s_T)]
% \label{eq:exp_potential_1}
% \end{equation}
%
In Equation \ref*{eq:exp_potential_3}, we simulate the "Go-phase" of Go-Explore over the world model $\hat{M}$. We set each state from $G_{edge}$ as the goal command $g$ for the goal-conditioned policy $\pi^G$ to run over $\hat{M}$ and denote $s_T$ as the final state of the resulting imagined trajectory %Please note that the $g$ sampled from the latent space needs to be transformed into the goal input dimensions set by $\pi^G$ firstly. 
(here $\hat{p}_{\pi^G(\cdot \vert \cdot,g)(\tau)}$ essentially induces the imagined trajectory distribution over the world model). %presented by the world model dynamic transition and goal-conditioned policy distributions.
In our implementation, we set the length of "Go-phase" $T$ to half of the maximum episode length for all environments. The time limits for both the Go and Explore phases during real environment exploration are also set to this value. We use the learned exploration value function $V^E$ of explorer $\pi^E$ to estimate the exploration value of $s_T^k$, the final state of $k$-th imagined trajectory. We average the estimated exploration potential over $K$ such imagined trajectories. 

At line~\ref{alg:CE2:ge} in Algorithm~\ref{alg:CE2}, after selecting the exploration target $g^E$ with the highest exploration potential $P^E$ from the latent cluster boundaries, we start the Go-Explore procedure in the real environment by executing the goal-conditioned policy $\pi^G$ to approach $g^E$ as closely as possible limited in $T$ timesteps, followed by launching the explore policy $\pi^E$ for exploration limited in $T_E$ timesteps. %This process constitutes the Go-Explore exploration strategy.
%Since the objective in Equation \ref*{eq:exp_potential_1} is not easily computable, as it relies on the final state distribution induced by the target-conditioned 
%policy $\pi^g$, which may rapidly change throughout the training process, it's crucial to use the latest estimates for better exploration. 
%We achieve this by leveraging the learned world model.

%After selecting the exploration target $g^E$ with the highest exploration potential $P^E$, we first approximate $g^E$ using the agent's goal-conditional policy, 
%and then utilize the agent's explore policy to explore unknown areas. 

%where the current capability of the agent can reach the optimal trajectory. 
%This ensures the effectiveness and efficiency of the exploration process. 

\subsection{The Main Algorithm}

%The clusters in latent space are initialized and trained from the data in the replay buffer, which determines the shape of the clustering space.

\begin{wrapfigure}[13]{r}{0.59\textwidth}
\vspace{-15pt}
\scalebox{0.8}{
\begin{minipage}{0.72\textwidth}
\begin{algorithm}[H]
\caption{The main training algorithm for \tool{}}
\label{alg:train_loop_exp}
\begin{algorithmic}[1]
\State \textbf{Input:} $\pi^G$, $\pi^E$, World Model $\hat{M}$, {\sc GMM}, $r^G$, $r^E$%, Clustering Centroids Number $N_{cluster}$
\State Initialize replay buffer $D$
\For{$i = 1$ to $N_{train}$}
    \If{Should assign centroids}
        \State $B_{exp} \leftarrow$ A batch of data from $D_{exp}$
        \State {\sc GMM} $\leftarrow$ Choose $N_{c}$ centroids from $B_{exp}$ by FPS
        \EndIf\label{alg:exp:reinit}
    \State $D_{exp} \leftarrow$ Cluster Edge Exploration(...) with Algorithm \ref{alg:CE2}\label{alg:exp:exp}
    \State $D \leftarrow D \cup D_{exp}$
    \State Update $\hat{M}$ with $D$ (update latent space by $\mathcal{L}_{rec}+\mathcal{L}_{latent}$)
    \State Update {\sc GMM} with $D_{exp}$
    \label{alg:ce2:updategmm}
    \State Update $\pi^G$ in imagination with $\hat{M}$ to maximize $r^G$
    \State Update $\pi^E$ in imagination with $\hat{M}$ to maximize $r^E$
    \EndFor
\end{algorithmic}
\end{algorithm}
\end{minipage}
}
\end{wrapfigure}

We depict the main learning algorithm of \tool{} in Algorithm~\ref{alg:train_loop_exp}. Recall that the learning objective is to train an agent that can achieve diverse goals revealed to it only at test time. Accordingly, in this algorithm at line~\ref{alg:exp:exp}, the data $D_{exp}$ collected to train the world model $\hat{M}$ is generated solely by our Go-Explore strategy as outlined in Algorithm~\ref{alg:CE2}. At line~\ref{alg:exp:reinit}, we periodically update the centroids of the latent clusters again using the FPS algorithm~(\cite{eldar1997farthest}) from a batch of latest trajectories from the replay buffer. %By utilizing the latest trajectory data from the replay buffer to train clusters, 
This ensures that the candidate goal states selected for exploration are indeed located at the boundaries of key state regions.
%This ensures that the clusters do not aggregate into a single large cluster during the training process. The Algorithm \ref*{alg:CE2} shows our method.
%To ensure that the GMM reflects the latest capabilities of the agent, 
At line~\ref{alg:ce2:updategmm}, we train the clustering model using data from the replay buffer in each round. This ensures that latent state clustering and the agent's goal-reaching capability are kept synchronized.

%This data structure of $D_{exp}$ determines that the clustering space will cover the entire environmental space as much as possible and encourages the agent to explore all unknown areas in environment sapce.

%\begin{wrapfigure}[11]{r}{0.55\textwidth}
%\vspace{-1cm}
\begin{wrapfigure}[15]{r}{0.59\textwidth}
\vspace{-15pt}
\scalebox{0.8}{
\begin{minipage}{0.72\textwidth}
\begin{algorithm}[H]
\caption{The main training algorithm for \toolegc{}}
\label{alg:train_loop_egc}
\begin{algorithmic}[1]
\State \textbf{Input:} $\pi^G$, $\pi^E$, $G$, World Model $\hat{M}$, {\sc GMM}, $r^G$, $r^E$, $p_g$%, Clustering Centroids Number $N_{cluster}$
\State Initialize replay buffer $D$
\For{$i = 1$ to $N_{train}$}
    \If{Should assign centroids}
        \State $B_{egc} \leftarrow$ A batch of data from $D_{egc}$
        \State $GMM \leftarrow$ Choose $N_{c}$ centroids from $B_{egc}$ by FPS
        \EndIf
    \State $D_{exp} \leftarrow$ Cluster Edge Exploration(...) with Algorithm \ref{alg:CE2}
    \State $D_{egc} \leftarrow$ Rollouts of $\pi^G$ using the env goal distribution $p_g$
    \State $D \leftarrow D \cup D_{exp} \cup D_{egc}$
    \State Update $\hat{M}$ with $D$ (update latent space by $\mathcal{L}_{rec}+\mathcal{L}_{latent}$)
    \State Update GMM with $D_{egc}$
    \State Update $\pi^G$ in imagination with $\hat{M}$ to maximize $r^G$
    \State Update $\pi^E$ in imagination with $\hat{M}$ to maximize $r^E$
    \EndFor
\end{algorithmic}
\end{algorithm}
\end{minipage}
}
\end{wrapfigure}

In our experiment, we also designed a variant of \tool{} in Algorithm~\ref{alg:train_loop_egc}, called \toolegc{}. This algorithm is given the environment goal distribution $p_g$ at training time. The main idea is to progressively expand the scope of exploration around the possible trajectories leading to the environment goals. 
In this algorithm, the replay buffer additionally includes $D_{egc}$ the trajectories sampled by $\pi_G$ conditioned on the environment goals in $p_g$. We only use $D_{egc}$ to initialize and train latent state clusters. 
In this way, the agent is encouraged to prioritize exploration starting from the edges of latent state clusters along the trajectories towards the goal states in $p_g$. \toolegc{} can be considered as learning policies and world models specific to a given goal distribution. 

\section{Experiments}

\begin{figure}[t] 
  \centering
  \includegraphics[width=0.9\textwidth]{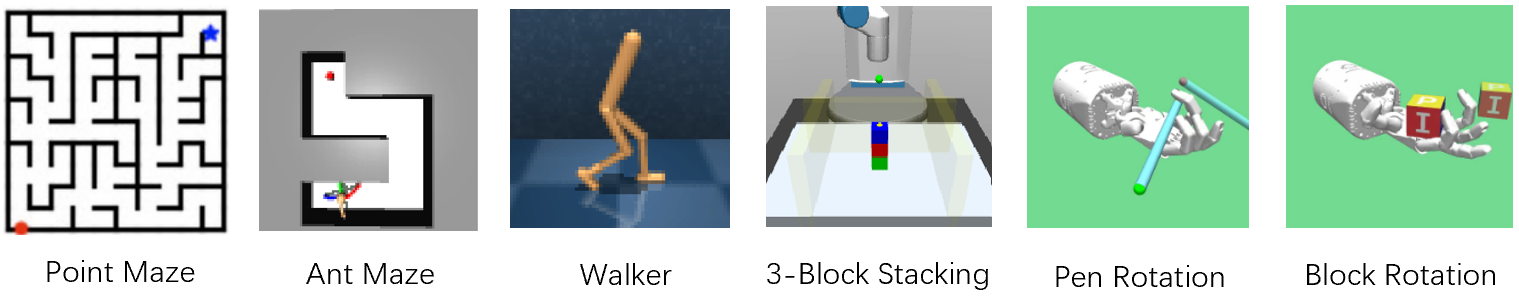}
  \caption{We conduct experiments on 6 environments: Point Maze, Ant Maze, Walker, 3-Block Stacking, Block Rotation, Pen Rotation.}
  \label{fig:environments}
\end{figure}

% In this section, we use a large number of experiments to prove that our method outperforms other exploration goal-picking strategy when the agent deal with a series of hard-exploration tasks. 
% We also visualize the process of exploration to see the essential difference of goal selection behaviors between \tool{} and other goal-picking strategies.
% Finally, we study the effect with different cluster number in the latent space to pick exploration goal from and how much is the performance gain brought by PEG in our method.

Our experiments evaluate \tool{} over goal-reaching tasks that demand significant exploration to solve. We aim to address the following questions: (1) Does \tool{} lead to improved exploration and goal-reaching performance? (2) How does \tool{} exploration qualitatively differ from those in previous goal-directed exploration methods? (3) Which components of \tool{} are crucial to its success?

\subsection{Benchmarks}
We evaluate our method on six hard exploration goal-conditioned RL tasks: \textbf{Point-Maze}, \textbf{Ant-Maze}, \textbf{Walker}, \textbf{3-Block Stacking}, \textbf{Block Rotation} and  \textbf{Pen Rotation}.
\textbf{Point-Maze}: A blue point is placed at the bottom left of the maze and be trained to explore the structure of maze.
\textbf{Ant-Maze}: An ant robot must master intricate four-legged locomotion behaviors and maneuver through narrow hallways. 
\textbf{Walker}: A 2-legged robot needs to learn how to control its leg joints to walk on a flat plane to move forward or backward.
In \textbf{3-Block Stacking}, a robot arm with a two-fingered gripper operates on a tabletop with three blocks. The goal is to stack the blocks into a tower configuration. The agent needs to learn pushing, picking, and stacking, as well as discovering intricate action paths to accomplish the task within the environment. Previous solutions have relied on methods like demonstrations, curriculum learning, or extensive simulator data, highlighting the task's difficulty.
The Gymnasium \textbf{Block Rotation} and \textbf{Pen Rotation} tasks involve manipulating a block and a pen, respectively, to achieve a random target rotation along all axes. %Block Rotation focuses on rotating a block with random target rotations , while 
Pen Rotation is particularly challenging due to the pen's thinness, requiring precise control to prevent it from dropping.
For evaluation, we use the most challenging goals, such as the farthest goal locations, in Point Maze, Ant Maze, Walker, and 3-Block Stacking. In the other two environments, we utilize random goals as defined by the environment.
For more settings and information about the environments, please refer to the Appendix~\ref{sec: envs}.

\subsection{Baselines}
%We tackle the exploration challenge of GCRL, focusing on how to plan efficient goals for exploration during training. There are two scenarios when goal-picking strategies are employed to expedite exploration: one involves not utilizing original goals provided by the environments, relying entirely on self-induced goals for training; the other involves utilizing both original goals from the environments and self-induced goals generated by goal-picking strategies simultaneously. As original goals returned by the environments can often be too challenging for the agent's abilities, or the goal-picking strategies may not require their utilization to induce goals for exploration, some exploration methods fall into the first scenario. We consider both scenarios and present two versions of our method: one is named \textbf{\tool{}}, and the other is \textbf{\toolegc{}}.
In the unsupervised GCRL setting, we compared \textbf{\tool{}} with state-of-the-art methods based on the Go-Explore strategy, which has demonstrated high efficiency in this setup:  \textbf{PEG}~(\cite{hu2023planning}) and \textbf{MEGA}~(\cite{pitis2020maximum})\footnote{Our model-based MEGA baseline is borrowed from~(\cite{hu2023planning})}. MEGA commands the agent to rarely seen states at the frontier by using kernel density estimates (KDE) of state densities and chooses low-density goals from the replay buffer. PEG selects goal commands to guide an agent's goal-conditioned policy toward states with the highest exploration potential given its current level of training. This potential is defined as the expected accumulated exploration reward during the Explore-phase.

In scenarios where environment goal distributions are available to the agents, we compare \toolegc{} with \textbf{GC-Dreamer} (illustrated in Sec.~\ref{sec:psb}), \textbf{PEG-G}, \textbf{MEGA-G} and \textbf{L3P}. Similar to \toolegc{}, PEG-G and MEGA-G augment \textbf{GC-Dreamer} with the PEG and MEGA Go-Explore strategies, respectively. In these methods, the replay buffer $D$ contains not only trajectories sampled by the goal-conditioned policy $\pi_G$ commanded by environment goals but also exploratory trajectories sampled using the corresponding Go-Explore strategies. \textbf{L3P} trains a latent space using temporal distances and performs clustering in this latent space, similar to \toolegc{}. %The trajectories to initialize and train the clusters are also targeted with goals from the environment. 
However, \textbf{L3P} does not employ a Go-Explore strategy. Instead, it constructs a directed graph with cluster centroids as nodes and utilizes online planning with graph search to determine subgoals for task execution.

%In \tool{}, the agent does not utilize any goals from the environment; the clusters are trained solely from trajectories using goals generated by our algorithm. This setup aligns with the original PEG paper, and thus we use the baseline \textbf{PEG} and \textbf{MEGA} for comparison, following the implementation provided in the PEG paper's repository.
%For \toolegc{}, our algorithm employs trajectories sampled with goals returned from both the environment and our goal-picking strategy to train the world model. Additionally, it utilizes the former to initialize and train the clusters. Therefore, \toolegc{} incorporates original goals returned by the environments, despite their potential difficulty in the initial stages of training. To compare \toolegc{} with baselines that also utilize original goals from the environments, we select \textbf{GC-Dreamer}, \textbf{PEG+egc}, and \textbf{L3P}.
%\textbf{GC-Dreamer} is the goal-conditional variant of Dreamer realized by LEXA(\cite{mendonca2021discovering}), which employs goals from the environment to sample trajectories and populate the replay buffer. Subsequently, the world model and policy are trained based on the framework depicted in Fig~\ref{fig:mbrl}.
%\textbf{PEG+egc} not only uses goals induced from the PEG strategy but also employs original goals returned by the environment to sample trajectories. Afterward, \textbf{PEG+egc} utilizes a replay buffer composed of these mixed trajectories to plan PEG exploration goals for Go-Explore.

\subsection{Results}

\begin{figure}[t] 
  \centering
  \subfigure[Point Maze]{\includegraphics[width=0.25\textwidth]{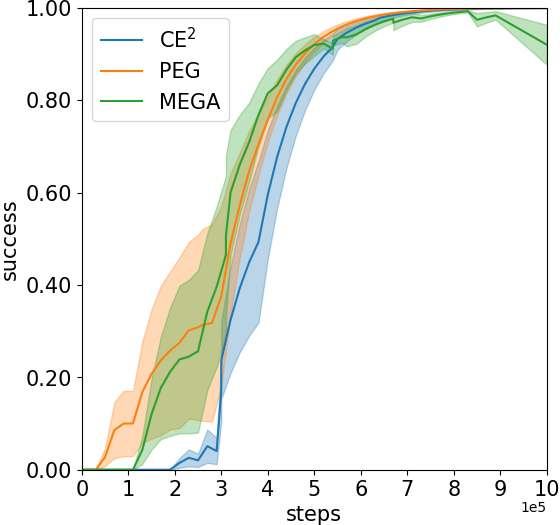}}
  \subfigure[Ant Maze]{\includegraphics[width=0.25\textwidth]{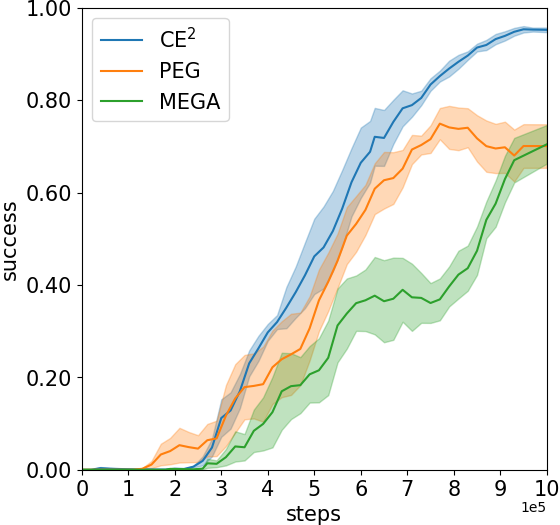}}
  \subfigure[Walker]{\includegraphics[width=0.25\textwidth]{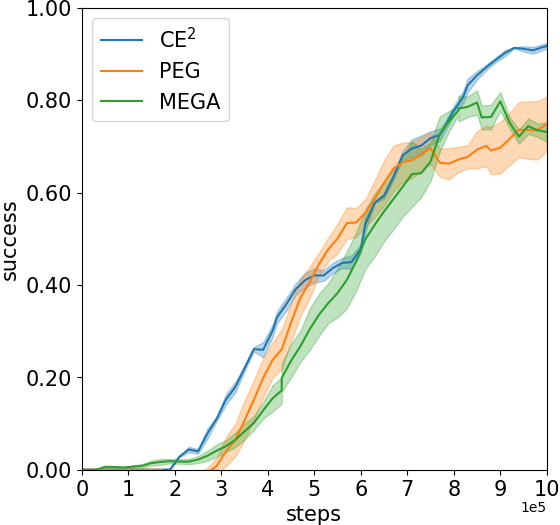}}
  \subfigure[3-Block Stacking]{\includegraphics[width=0.25\textwidth]{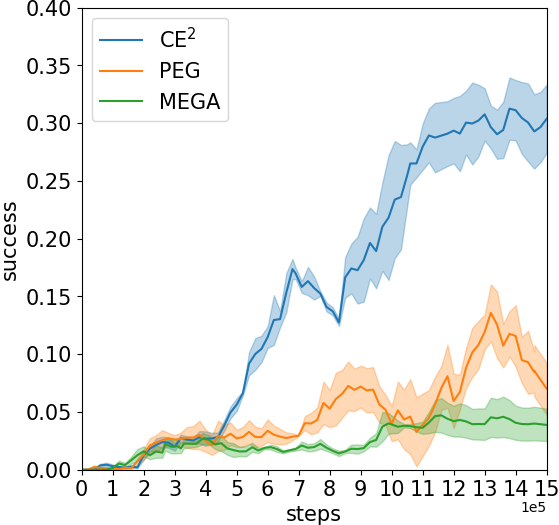}}
  \subfigure[Block Rotation]{\includegraphics[width=0.25\textwidth]{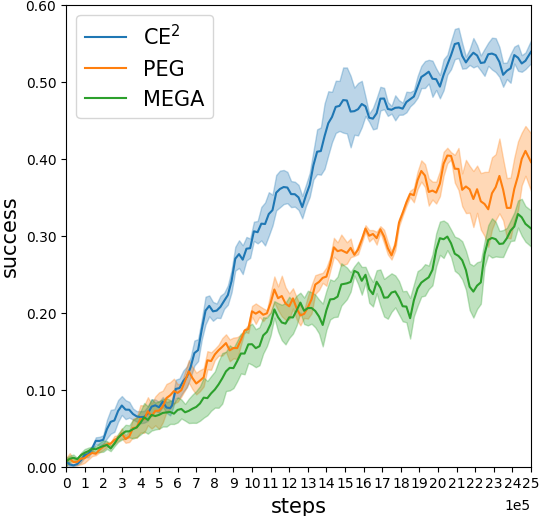}}
  \subfigure[Pen Rotation]{\includegraphics[width=0.25\textwidth]{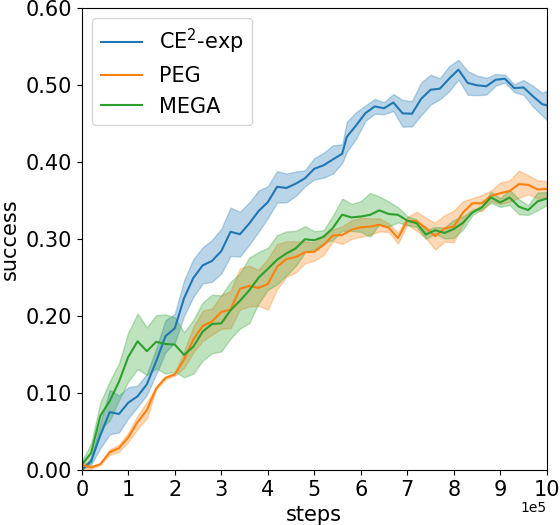}}
  \caption{Experiment results comparing \tool{} with the baselines over 5 random seeds.}
  \label{fig:exp_res_wo_goal}
\end{figure}

\begin{wrapfigure}[11]{r}{0.5\textwidth}
\vspace{-10pt}
  \centering
  \includegraphics[width=0.5\textwidth]{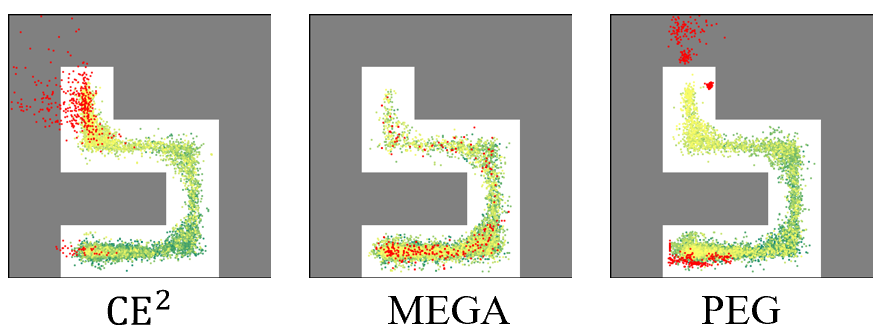}
  \vspace{-0.5cm}
  \caption{Comparison of exploration goals (represented as red points) generated by \tool{}, MEGA, and PEG in the Ant Maze environment.}
  \label{fig:ant-comparasion}
\end{wrapfigure}

\textbf{\tool{} Results.}
% In the first scenario where original goals from the environment are not utilized, we repeat \tool{}, 
% PEG and MEGA training respectively on the six environments for 5 trials with different random seeds for 1e6 timesteps 
% and report the average success rate across training time in Fig~\ref{fig:exp_res_wo_goal}.
Fig.~\ref{fig:exp_res_wo_goal} depicts the mean learning performance of all the unsupervised GCRL tools in terms of the agent's goal-reaching success rate averaged over 5 random seeds. The evaluation goal distribution is revealed to the agent only at test time. In all tasks except PointMaze, \tool{} significantly outperforms PEG and MEGA in terms of both learning performance and learning speed. On PointMaze, \tool{} performs comparably with the baselines.
%For 3 Block, Ant Maze,  Block Rotation and Pen Rotation environments, 
%While in Point Maze and Walker environments, \tool{} shows competitive performance compared with PEG and MEGA.
%We specifically pay attention to environments requiring intricate joint control, such as rotating objects and stacking three blocks. 
%In these environments, \tool{} demonstrates a significant advantage over the most powerful and state-of-the-art exploration strategies, PEG and MEGA. 
Although MEGA can set goal commands in sparsely explored areas of the state space to encourage exploration, unlike \tool{}, it lacks a systematic method to filter out unachievable goals for the agent, which can result in inefficient exploration. Theoretically, PEG can induce more exploration than MEGA because it can sample goal commands as any state within the state space to initiate exploration, including those beyond the frontier of known states in the replay buffer. However, because a learned world model is typically unfamiliar with rarely observed states, it may select goal commands that appear to have high exploration potential in the model but perform poorly in the real environment as shown in Fig.~\ref{fig:ant-comparasion}.

%PEG  utilizes world model to predict exploration potential, and MEGA selects low-density targets from the replay buffer. 
%This illustrates that \tool{} is not only capable of sampling low-density targets reachable by the agent in edge of latent space clusters 
%but also of filtering out targets with higher exploration value through predicted exploration potential. 
%These two components of \tool{} respectively embody the core ideas of MEGA and PEG, thereby showcasing significant results in simultaneously outperforming both strategies.

\begin{figure}[t] 
  \centering
    \subfigure[Walker]{\includegraphics[width=0.24\textwidth]{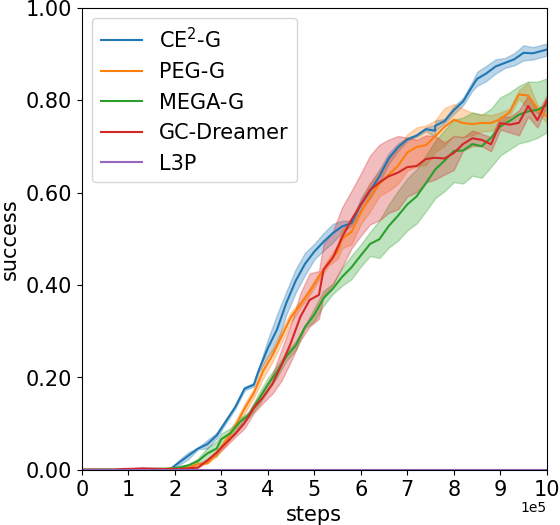}}
  \subfigure[3-Block Stacking]{\includegraphics[width=0.24\textwidth]{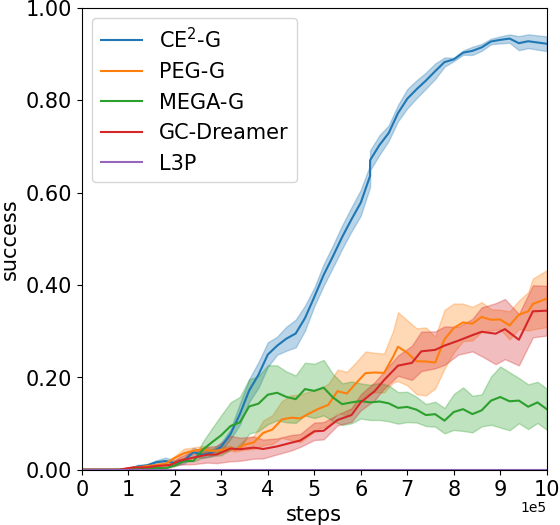}}
  \subfigure[Block Rotation]{\includegraphics[width=0.24\textwidth]{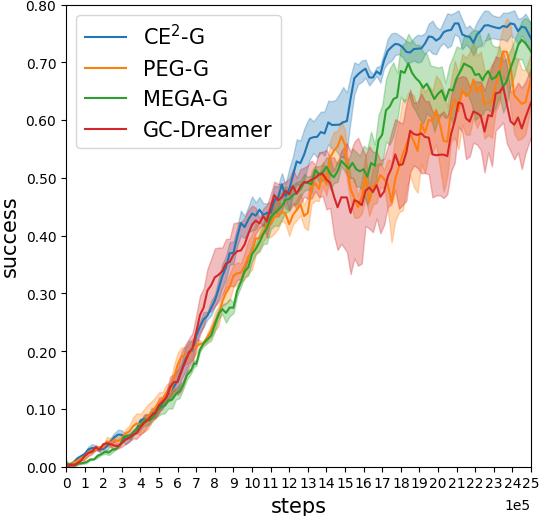}}
  \subfigure[Pen Rotation]{\includegraphics[width=0.24\textwidth]{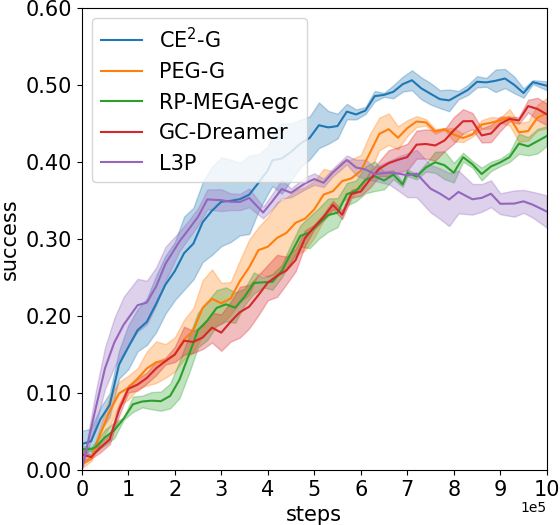}}
  \caption{Experiment results comparing \toolegc{} with the baselines over 5 random seeds.}
  \label{fig:exp_res_goal}
\end{figure}

\textbf{\toolegc{} Results.} 
Fig.~\ref{fig:exp_res_goal} depicts the mean learning performance of all the tools in terms of the agent's goal-reaching success rate averaged over 5 random seeds when the environment goal distribution is revealed to the agent at training time. 
GC-Dreamer is the only tool that lacks a Go-Explore phase, which may limit its exploration potential. Even so, it can sometimes outperform MEGA-G and PEG-G (see block rotation and pen rotation). This indicates that, without reasonably accounting the agent's capability to reach selected goal commands, the Go-Explore strategy does not always guarantee improved exploration. Suboptimal goal-setting during the "Go-phase" can even hinder exploration (see 3 block stacking). Notably, for the challenging 3-block stacking task, \toolegc{} achieves a high success rate exceeding 90\%. In comparison, MEGA-G, PEG-G and GC-Dreamer only achieve less than 40\% success rates. Refer to Appendix~\ref{subs: fullresultforegc} for full results of \toolegc{}.

\subsection{Exploration Process}

\begin{figure}[t] 
  \centering
  \includegraphics[width=1\textwidth]{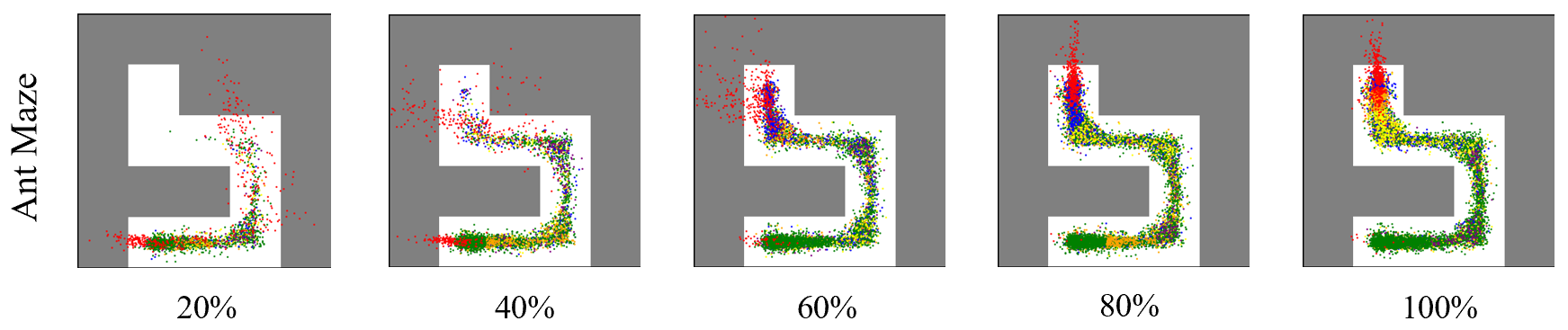}
  \caption{Cluster evolution in \tool{} as the training progresses. The red points means the goals picked by \tool{} to explore and other points in different colors represent the clusters \tool{} learned.}
  \label{fig:cluster visualization}
\end{figure}

Fig~\ref{fig:cluster visualization} shows the evolution of state clusters (learned in a latent space) during the training process for Ant Maze (in different colors). The red points represent the selected goal commands used to induce exploration. %It is shown that \tool{} effectively guides the agent to states at the cluster edges, which hold high exploratory potential. 
We observe that the self-directed exploration goals set by \tool{} improve progressively as the agent's capabilities increase, consistently targeting the cluster edges that require further exploration and are within the agent's reach. We compare the exploration targets generated by \tool{} with those produced by the MEGA and PEG approaches throughout the training process in the Ant Maze environment in Appendix \ref{subs: MoreAntExploreSpace}.
%It is noteworthy that clustering is not directly conducted in the original state space, but rather in a latent space trained with a temporal distance network. 
%Hence, our clustering appears intertwined and overlapping in the state space, especially during the early stages of training when the temporal distance network $D_t$ is not yet well-trained.
%We sample at the edges of clusters in the latent space, thereby exploring areas with limited agent knowledge in a constrained manner.

\subsection{Ablation Study}

\begin{figure}[t] 
  \centering
    \subfigure[Ant Maze]{\includegraphics[width=0.24\textwidth]{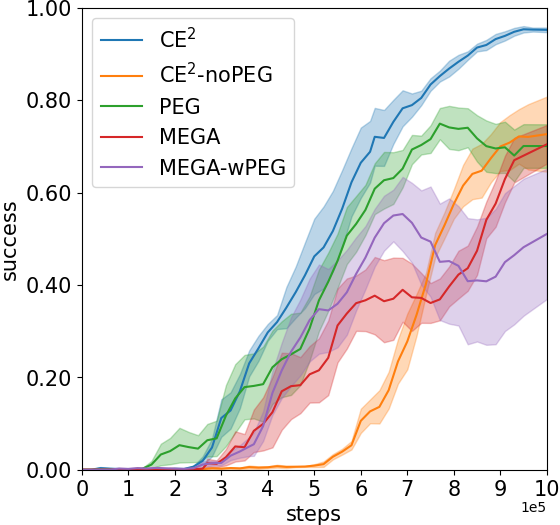}}
  \subfigure[3-Block Stacking]{\includegraphics[width=0.24\textwidth]{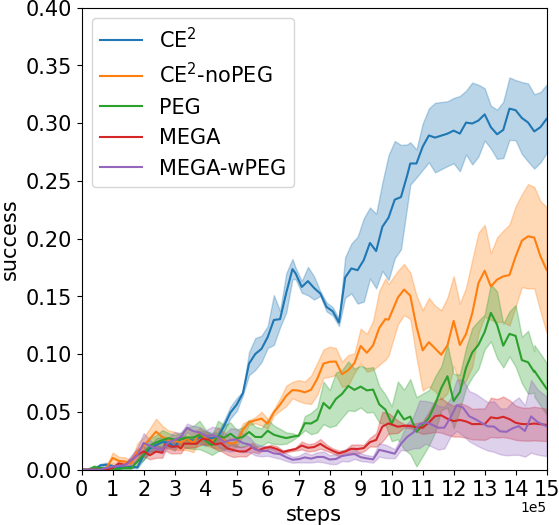}}
  \subfigure[Block Rotation]{\includegraphics[width=0.24\textwidth]{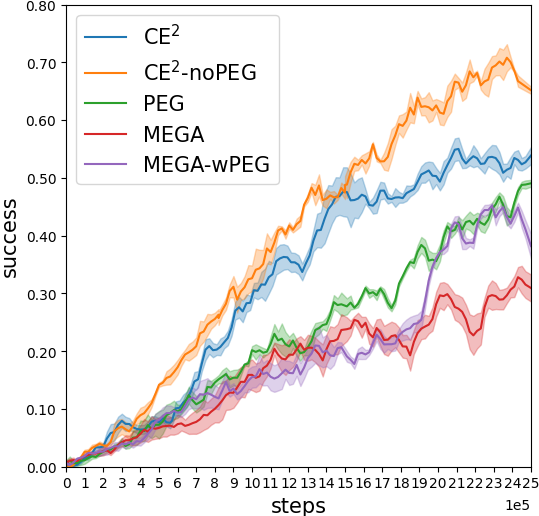}}
    \subfigure[Pen Rotation]{\includegraphics[width=0.24\textwidth]{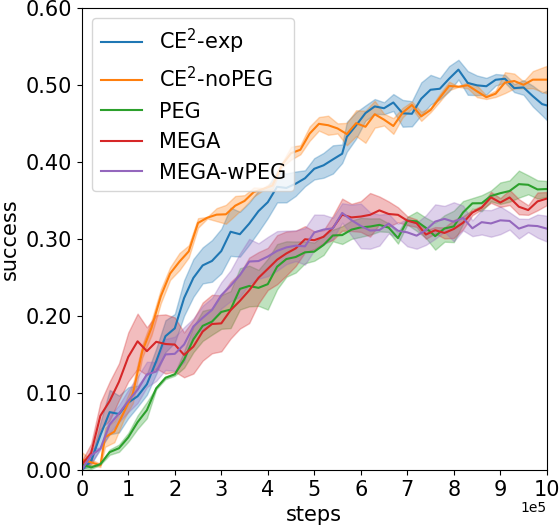}}
  \caption{Ablation study on the importance of each component of \tool{} over 5 random seeds.}
  \label{fig:exp_abl_1}
\end{figure}

In the ablation experiment, our goal is to determine the individual contributions of each component to our method's overall performance. The "Go-phase" of the Go-Explore procedure in \tool{} consists of two main steps for selecting a goal command $g$ to initiate exploration: (a) sampling environment states at the boundaries of its trained latent state clusters, and (b) selecting the goal command $g$ with the highest exploration potential from the sampled states. Our first ablation, \textbf{\toolwithoutpeg{}}, only performs step (a). It randomly samples $g$ from the latent state clusters without considering its exploration potential. The second ablation only performs step (b) and is identical to the PEG baseline. It can sample any state within the state space for the goal command $g$, without the constraint of directing the agent to states at the boundaries of known regions like \tool{} and MEGA. 
We also include \textbf{MEGA} and \textbf{MEGA-wPEG} as two baselines to solely compare the exploration strategy-step (a)-in \tool{} with MEGA's strategy to command the agent to rarely seen states. MEGA-wPEG first uses MEGA to sample a batch of candidate goals, all with low density in the replay buffer. Then, their exploration potential is evaluated using PEG (step (b)), and the most valuable one is selected as the exploratory goal. We conduct the ablation experiments in a purely unsupervised setting without revealing any test goals at training time.

%These two baselines provide another goal-directed exploration strategy to command the agent to rarely seen states.

%parts to pick a goal for exploration.  The first one is to sample candidate goals from the edge of clusters, the second is to evaluate the exploration potential of these candidate goals using world model and pick out the goal with maximum exploration potential.
%So, we want to know which of these two parts contributes more to our success. 

%If we retain the first part and remove the second part, that is, we randomly select a point from the candidate points sampled at the edges of cluster as the exploration goal, 
%we refer to this modified version as \textbf{\toolwithoutpeg{}}. 
%\textbf{MEGA} uses kernel density estimates (KDE) of state densities and selects frontier goals with low density from the replay buffer. 
%To further strengthen the evaluation of the role of the second part, we conduct experiments for \textbf{MEGA-wPEG}. 
%\textbf{MEGA-wPEG} firstly employs MEGA to sample a batch of candidate goals, all of which have low density in the replay buffer. 
%Subsequently, their exploration potential is evaluated using PEG, with the most valuable one selected as the exploration goal.
%We can get the importance of first part by comparing \textbf{MEGA-wPEG} and \textbf{\tool{}} or comparing \textbf{MEGA} and \textbf{\toolwithoutpeg{}} or comparing \textbf{PEG} and \textbf{\tool{}}
%We can also know the role of second part by comparing \textbf{MEGA} and \textbf{MEGA-wPEG} or comparing \textbf{\toolwithoutpeg{}} and \textbf{\tool{}}. 

Fig.~\ref{fig:exp_abl_1} confirms that both step (a) and step (b) in \tool{} are important. \tool{} significantly outperforms \toolwithoutpeg{} and PEG in 3-block stacking and the Ant maze tasks. Notably, even without step (b), \toolwithoutpeg{} performs well across all experiments, especially in the challenging block and pen rotation tasks. This indicates that the goal commands sampled at the edges of latent state clusters already possess high exploration potential and can guide the agent to traverse unseen state spaces. 
The superior performance of \toolwithoutpeg{} compared to both MEGA and MEGA-wPEG further reinforces this. Block Rotation is the only environment where \toolwithoutpeg{} outperforms \tool{}. %In \tool{}, our method selects goals by identifying states with the highest exploration value estimate sampled from the latent state cluster boundaries to initiate our Go-Explore procedure. Since we consider unsupervised exploration, the test goal distribution is not available to the \tool{} agent during training. 
In this environment, the \tool{} agent often pursues states where the block falls from the palm, due to their "high" exploration potential determined by the exploration policy value functions. In contrast, \toolwithoutpeg{} agent explores the state space more evenly, gaining more in-hand manipulation skills, which is crucial for achieving the block-rotation goals revealed at test time. MEGA achieves similar or better performance compared to MEGA-wPEG, indicating that PEG's effectiveness relies on the quality of the candidate goal set. The exploratory goals sampled from the lowest-density regions in the replay buffer might be beyond the agent's capability, leading PEG to assess the true exploration potential of the candidate goals inaccurately.

%Fig~\ref{fig:exp_abl_1} show our results on three environments. \tool{} outperforms PEG in final success rate and learning speed of all three environments which proves that candidate goals sampling plays a nontrivial role to remove noisy goals for further selection in second part and accelerates policy training. 
%Moreover, based on the results that \tool{} achieves better performance than MEGA+PEG and \toolwithoutpeg{} gets higher success rate than MEGA, we can conclude that candidate goals sampled from cluster edge are more valuable and essential for exploration compared with that sampled based on density. 
%To evaluate the second part of \tool{}, \tool{} achieves higher success rate compared with \toolwithoutpeg{}, which indicates the importance of world model evaluation in our method. However, 

We also conducted experiments in the \tool{} with different numbers of latent state clusters $N_c$ and observed that \tool{} is insensitive to this hyperparameter. See Appendix~\ref{subs: moreablation} for more discussion.

%to observe if the results are sensitive to the number of clusters. 

\section{Related Work}

Our method addresses the challenging and inefficient exploration problem inherent in goal-conditioned reinforcement learning (RL) settings with sparse rewards, 
commonly used in robotics and control fields (\cite{ghosh2019learning, liu2022goal,plappert2018multi}). 
In goal-conditioned RL, agents are trained to achieve various goals based on predefined commands, with rewards typically being binary, 
indicating positive feedback from the environment only upon reaching the specified goal. 
This sparse reward setting significantly complicates achieving sample efficiency and effective 
learning processes (\cite{ren2019exploration, florensa2018automatic, trott2019keeping}). 
To mitigate this challenge, various methods have been proposed. Some reshape the sparse reward function into a denser form by incorporating 
metrics such as distance between achieved and desired goals(\cite{trott2019keeping}) or temporal distance (\cite{hartikainen2019dynamical,mendonca2021discovering}). 
Additionally, exploration strategies often include rewards aimed at incentivizing visits to states with low visitation 
frequencies (\cite{bellemare2016unifying, burda2018exploration}). These approaches typically involve identifying states 
with infrequent occurrences within the replay buffer and targeting them for exploration, thus facilitating the discovery of unknown regions 
in the environment. Furthermore, some research emphasizes the exploration of states with high variance between ensemble predictions 
of future states (\cite{mccarthy2021imaginary, oudeyer2007intrinsic, pathak2017curiosity, henaff2019explicit,shyam2019model, sekar2020planning}).

In addition to reshaping the exploration reward function, goal-directed exploration represents a widely employed strategy that sets exploration goals
distinct from the final task objective. Essentially, this approach aims to select goals that present challenges to the current policy while 
remaining achievable. Prior works have proposed various methods to generate goals for goal-directed exploration.
(\cite{zhang2020automatic}) proposed to do automatic curriculum generation of goals based on the epistemic uncertainty of value functions.
(\cite{florensa2018automatic}) use generative adversarial training to automatically generate goals, leveraging goal difficulty as a guiding factor.
(\cite{pong2019skew, pitis2020maximum}) proposed to use the maximum entropy of achieved goal distribution to guide goal selection.
(\cite{ecoffet2019go}) introduce a more efficient exploration methodology known as Go-Explore. This approach initially employs the goal-conditioned policy (Go-phase), 
followed by the rollout of the exploration policy from the terminal state of the goal-conditioned phase (Explore-phase). Go-Explore facilitates 
exploration initiation from a state area accessible by the current capabilities of the goal-conditioned policy.

PEG (\cite{hu2023planning}) proposes computing the exploration potential by simulating Go-Explore 
trajectories using a world model to identify goals characterized by elevated average exploration rewards in the Explore-phase. 
This metric incorporates anticipated exploration rewards of the Explore-phase, providing an advantage for Go-Explore. However, the goals sampled 
for evaluating this exploration potential metric in PEG are drawn from a distribution updated by the MPPI method (\cite{williams2015model,nagabandi2020deep}) directly in the 
observation space. L3P (\cite{zhang2021world}) employs temporal distance to train a latent space, facilitating clustering within this space to delineate key 
state areas based on reachability. Our approach proposes exploration from the periphery of these key state regions, aiming to balance exploration of 
unknown territories while constraining exploration starting points to the edges of key state regions, thus avoiding meaningless exploration from widely 
sampled points from observation space. See Appendix~\ref{sec: discussion}, ~\ref{sec: morerelatedwork} for more related work discussion.

\section{Conclusion}

We present \tool{}, a novel Go-Explore mechanism designed to tackle hard exploration problems in unsupervised goal-conditioned reinforcement learning tasks. %During the "Go-phase", \tool{} %selects exploration-inducing goals by explicitly considering the agent's current capabilities. This approach 
%efficiently guides the agent to the frontier of known states, represented as the boundaries of clusters of states in the replay buffer that are easily reachable from one another. %We instantiate \tool{} within the context of model-based GCRL, demonstrating how learned world models can cluster environment states based on their temporal distance in a latent space. 
While \tool{} outperforms prior exploration approaches in challenging robotics scenarios, the requirement to learn state clusters to identify frontier states and the reliance on world models to determine exploration potential introduce nontrivial computational costs. Exploring whether \tool{}'s Go-Explore strategy can be effectively applied to model-free GCRL settings remains an interesting avenue for future work.

%We validate \tool{}'s effectiveness in challenging robotics scenarios, showing that it leads to more efficient and effective training of adaptable GCRL policies compared to baseline methods and ablations.

%Compared to PEG which is a purely unsupervised exploration approaches, 
%we have two version of \tool{}, distinguished by whether they utilize task goal space. 
%One is \tool{} which is also a totally unsupervised exploration approaches, autonomously planing to explore the environment as extensively as possible.
%The other is \toolegc{} which enables exploration around the path leading to the ultimate goal region of task, 
%facilitating more efficient exploration of task-oriented environmental spaces. 
% \tool{} has the capability to leverage the current abilities of the agent to perform clustering in the Latent Space, identifying crucial state regions. 
% Through sampling at the edges of clusters, it can effectively seek points that provide exploration gains in a constrained manner, thus preventing aimless exploration. 
% Despite \tool{} still faces challenges with some complex state spaces and learning clusters in such environments requires more time and well learned latent space,
% Our method is inspiring for the development of an effective exploration approach that is constrained and task-oriented, thereby enhancing the learning ability of agents in challenging tasks.

\section*{Reproducibility Statement} 

The codebase of our method is provided on \hyperlink{https://github.com/RU-Automated-Reasoning-Group/CE2}{https://github.com/RU-Automated-Reasoning-Group/CE2}. For hyperparameter settings and baselines' pesudocode, please refer to Appendix~\ref{supp:baselines} and ~\ref{subsec: hyperparameters}.

\section*{Acknowledgements}

We thank the anonymous reviewers for their comments and suggestions. This work was supported by NSF Award \#CCF-2124155.

\bibliography{Reference/Reference}
\bibliographystyle{apalike}

%%%%%%%%%%%%%%%%%%%%%%%%%%%%%%%%%%%%%%%%%%%%%%%%%%%%%%%%%%%%

% Start the appendix

\newpage 

\appendix
\addcontentsline{toc}{section}{Appendix} 
{\fontsize{20}{14}\selectfont \textbf{Appendix}}
\parttoc 

\section{Discussion}
\label{sec: discussion}

\textbf{Why does \tool{} perform better than the original Go-Explore mechanism?}

Our algorithm, \tool{}, tackles the core challenge in the Go-Explore mechanism: how to select an exploration-inducing goal command $g$ and effectively guide the agent to $g$? Previous approaches, such as MEGA, set exploratory goals at rarely visited regions of the state space. However, in these approaches, the policies under training may have limited capability of reaching the chosen rare goals, leading to less effective exploration. Our contribution is a novel goal selection algorithm that prioritizes goal states in sparsely explored areas of the state space, provided they remain accessible to the agent. This is the key factor in why \tool{} outperforms the MEGA and PEG baselines in our benchmark suite in Fig.~\ref{fig:exp_res_wo_goal}. As visualized in Fig.~\ref{fig:MoreAntExploreSpace} in the appendix for the Ant Maze environment, \tool{} enhances exploration efficiency by consistently setting exploratory goals within the current policy's capabilities. In contrast, MEGA and PEG often set goals that are unlikely to be reachable by the current agent.
 
\textbf{Why don't we choose the original Go-Explore as a direct baseline?}

As discussed above, the core challenge in the Go-Explore mechanism lies in selecting goal states that effectively trigger further exploration upon being reached. However, the original Go-Explore method (\cite{ecoffet2019go}) does not prescribe a general goal selection method, instead opting for a hand-engineered novelty bonus for each task (e.g. task-specific pseudo-count tables). \tool{}
 is more related to recent instantiations of Go-Explore that automatically selects exploration-inducing goals in less-visited areas of the state space to broaden the range of reachable states, e.g. MEGA and PEG. Therefore, we compare our method with these tools instead of \cite{ecoffet2019go} in environments where these tools are applicable, to evaluate the strength of our goal selection method.

\textbf{Why our clustering algorithm does not structure the latent space in the learning process?}

While our clustering algorithm does not directly structure the latent space, it requires the latent space to be organized in a specific manner to be effective. 
In other words, the latent space learning algorithm is a key prerequisite for the latent state clustering algorithm. Specifically, our latent space learning algorithm structures the latent space such that states easily reachable from one another in the real environment (as determined by the learned temporal distance network as Equation~\ref{eq: temporal_distance_reward}) are also close together in the latent space. The clustering algorithm leverages this structure-property to ensure that the latent state cluster boundaries align with the frontier of previously explored states. As such, \tool{} can efficiently generate exploratory goals at the frontier at training time.

\section{Extended Related Work}
\label{sec: morerelatedwork}

Model-based reinforcement learning (MBRL) has seen significant advancements in recent years, driven by the development of sophisticated world models and planning algorithms. 
One notable approach is Stochastic Ensemble Value Expansion (STEVE) (\cite{buckman2018sample}), which enhances sample efficiency by leveraging ensemble models to reduce overfitting and uncertainty in value estimates. Similarly, the work by Chua et al. (\cite{chua2018deep}) demonstrates that probabilistic dynamics models can be effectively used to achieve high performance in a small number of trials.
In the realm of combining model-based and model-free methods, Deisenroth and Rasmussen (\cite{deisenroth2011pilco}) introduced PILCO, a data-efficient policy search method that uses Gaussian processes for dynamics modeling. More recent advancements include the integration of large pre-trained models for world model construction and task planning, as explored by Guan et al. (\cite{guan2023leveraging}).
The Dreamer framework by Hafner et al. (\cite{hafner2019dream}) utilizes latent imagination to learn behaviors directly from pixel observations, and its extensions (\cite{hafner2020mastering,hafner2023mastering}) have shown impressive results in mastering diverse domains. The Recurrent World Models by Ha and Schmidhuber (\cite{ha2018recurrent}) also contribute to this line of work by facilitating policy evolution through latent space planning.
Several approaches focus on improving exploration strategies within MBRL. For instance, the use of cross-entropy methods for Monte-Carlo Tree Search (\cite{chaslot2008cross}) and the Curious Replay mechanism (\cite{kauvar2023curious}) have been proposed to enhance exploration efficiency. The work by Wagenmaker et al. (\cite{wagenmaker2024optimal}) further explores optimal exploration strategies in nonlinear systems.
Additionally, transformers have been leveraged for their sample efficiency in world modeling (\cite{micheli2022transformers,zhang2024storm}), demonstrating their potential in complex environments. The integration of demonstrations into visual model-based reinforcement learning, as seen in MoDem (\cite{hansen2022modem}), showcases another avenue for improving learning efficiency.
Luo et al. (\cite{luo2018algorithmic}) provide a comprehensive framework with theoretical guarantees, while Janner et al. (\cite{janner2019trust}) address the critical question of when to trust the learned models.

\section{Extended Background}

\subsection{Dreamer World Model}
\label{subs: rssm}

We use the world model structure $\hat{M}$ of recurrent state-space model (RSSM) of Dreamer(\cite{hafner2019dream, hafner2019learning, hafner2020mastering, hafner2023mastering}) to learn the dynamics.
The complete model state of the RSSM is the concatenation of deterministic states and stochastic states, with the latter being generated by the former.
The deterministic state $h_t$ can used to get the prior state $\hat{z}_t$ and posterior state $z_t$. 
The $\hat{z}_t$ aims to predict the posterior without access to the current input state $x_t$ while the posterior state $z_t$ is concluded by integrating the encoded information of current input state $x_t$.
The deterministic state $h_t$ is updated by the recurrent transition function $f_{\phi}$ using the concatenation $(h_{t}, z_{t})$ or $(h_{t}, \hat{z}_t)$ as input. 
The world model is summarized in Fig~\ref{fig:world_model}, and the formulas of components are shown in Equation~\ref{eq:world_model}:

\begin{figure}[h]
  \centering
  \includegraphics[width=0.7\textwidth]{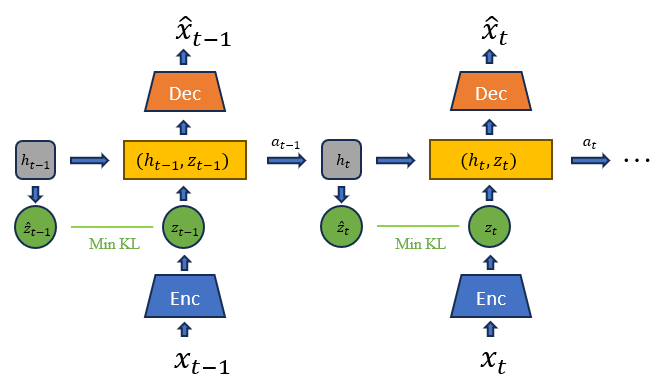}
  \caption{RSSM Structure}
  \label{fig:world_model}
\end{figure}

\begin{equation}\label{eq:world_model}
  \begin{aligned}
    \text{Encoder:\qquad} & e_t = f_E(e_t | x_t)\\
    \text{Recurrent model:\qquad} & h_t = f_\phi(h_{t-1}, z_{t-1}, a_{t-1}) \\
    \text{Representation model:\qquad} & z_t \sim q_\phi(z_t | h_t, e_t) \\
    \text{Transition predictor:\qquad} & \hat{z}_t \sim p_\phi(\hat{z}_t | h_t) \\
    \text{Decoder:\qquad} & \hat{x}_t \sim f_D(\hat{x}_t | h_t, z_t)
  \end{aligned}
\end{equation}

\subsection{Temporal Distance Training in LEXA}
\label{subs: Dt-training}

The goal-reaching reward $r^G$ is defined by the self-supervised temporal distance objective (\cite{mendonca2021discovering}) 
which aims to minimize the number of action steps needed to transition from the current state to a goal state within imagined rollouts.
We use $b_t$ to denote the concatenate of the deterministic state $h_t$ and the posterior state $z_t$ at time step $t$.

\begin{equation}\label{eq: temporal_distance_reward1}
  b_t = (h_t, z_t)
\end{equation}

The temporal distance $D_t$ is trained by sampling pairs of imagined states $b_t, b_{t+k}$ from imagined rollouts and predicting the action steps number $k$ between the embedding of them, 
with a predicted embedding $\hat{e}_t$ from $b_t$ to approximate the true embedding $e_t$ of the observation $x_t$.

\begin{equation}\label{eq: temporal_distance_reward2}
  \text{Predicted embedding:\qquad} emb(b_t) = \hat{e}_t \approx e_t, \qquad \text{where\quad} e_t = f_E(x_t)
\end{equation}

\begin{equation}\label{eq: temporal_distance_reward3}
  \text{Temporal distance: } D_t(\hat{e}_t, \hat{e}_{t+k}) \approx k/H \qquad \text{where\quad} \hat{e}_t = emb(b_t)\quad \hat{e}_{t+k} = emb(b_{t+k})
\end{equation}

\begin{equation}\label{eq: temporal_distance_reward4}
  r^G_t(b_t, b_{t+k}) = -D_t(\hat{e}_t, \hat{e}_{t+k})
\end{equation}

\section{Limitations and Future Work}

\begin{figure}
  \centering
  \includegraphics[width=1\textwidth]{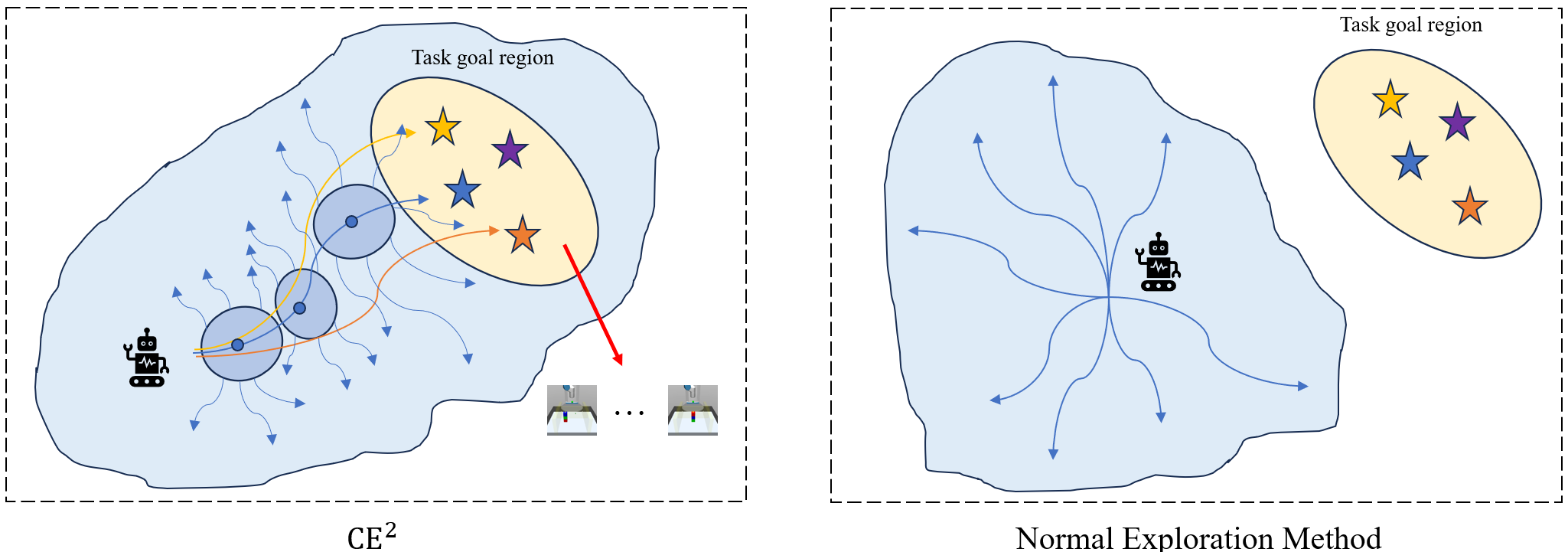}
  \caption{Illustration of differences between our mothod \toolegc{} and other exploration methods.}
  \label{fig:IllustrateDifferences}
\end{figure}

Our method clusters in the latent space, which necessitates a well-trained latent space. 
This latent space must not only accurately reconstruct the original state space and facilitate dynamic prediction but also reflect the reachable distances between different states. 
Therefore, training this latent space requires a temporal distance predictor that can accurately estimate the number of action steps needed between two states. 
We utilize the temporal distance predictor network from LEXA, which constructs intrinsic goal-conditioned rewards, and this network is trained using simulated trajectories. 
Compared to training the temporal distance predictor with real trajectories, using simulated trajectories offers greater stability. 
Our method requires the temporal distance predictor to reliably estimate the number of action steps needed to transition from one state to another, which is a crucial prerequisite for ensuring the effectiveness of \tool{}.
Moreover, Although \tool{} has achieved remarkable success in tasks such as stacking blocks and rotating objects, there remain more challenging tasks that \tool{} needs to address. 
For instance, environments such as inserting a peg into a hole or fluid tasks in ManiSkill2(\cite{gu2023maniskill2}).

Besides, our realization of \tool{} is based on Dreamer, a model-based reinforcement learning(MBRL) agent known for its higher sample efficiency but greater computational demands compared to model-free alternatives. 
This increased resource requirement stems from the necessity to develop a world model. 
In \tool{}, this world model is utilized to train policies and value functions through simulated trajectories.
At the same time, \tool{} use the PEG as the filter of exploration potential, which rely on world model to select goals that guide exploration. 
Creating a model-free version of \tool{} would simplify both its computational and conceptual aspects, a task we plan to undertake in future research.

\section{Environments}
\label{sec: envs}

\subsection{3-Block Stacking} 
In this task, a robot is required to stack three blocks into a tower. 
In PEG, evaluations are conducted on goals of varying difficulty levels, including 3 easy goals(picking up a single block), 6 medium goals(stacking two blocks), and 6 hard goals(stacking three blocks). 
We evaluate our agent solely on the 6 hard goals. At the same time, we use only 3 hard goals provided by the training environment as guiding goals for \toolegc{}.
Relying solely on the most challenging goals for training and evaluation presents a heightened challenge for both \tool{} and \toolegc{}. 
However, we observed that \tool{} and \toolegc{} are capable of spontaneously discovering additional easy and medium difficulty goals through clustering in latent space, as these serve as crucial transitional states towards the hard goals.
The environment features a 14-dimensional state and goal space: the first five dimensions capture the gripper's state, while the remaining nine dimensions correspond to the $xyz$ positions of each block. The action space is 4-dimensional, with three dimensions dedicated to the $xyz$ movements of the gripper and the fourth dimension controlling the gripper's finger movement. The robot achieves success when the L2 distance between each block's $xyz$ position and its target position is less than 3 cm. This environment is a modified version of the FetchStack3 environment from \cite{pitis2020maximum}, incorporating adjustments to better test the robot's precision in stacking.

\subsection{Walker} In this environment, a 2D walker robot is trained and evaluated the locomotion capabilities of  on a flat surface. The environment code is sourced from \cite{mendonca2021discovering}.
In order to fully evaluate the agent's ability and accuracy to travel to longer distances, 
we increased the number of evaluation goals in PEG from 4($\pm7, \pm12$) to 12($\pm13, \pm16, \pm19, \pm22, \pm25, \pm28$) along the $x$ axis from its initial position.
Noting that, in \toolegc{}, we only use goals at $\pm13, \pm16$ as the training goals returned by environments, but evaluate on all 12 goals.
Success is determined by checking if the agent's $x$ position falls within a small margin of the target $x$ position. 
The state and goal space are nine-dimensional, encompassing the walker's $xz$ positions and joint angles.

\subsection{Ant Maze} This environment is adapted from the Ant Maze described in \cite{pitis2020maximum}, with a little modifications. 
Like PEG, we set the goal space to be same with state space which including the ant's $xyz$ positions along with joint positions and velocities, and an additional room was added in the top left to introduce a more challenging goal.
In this complex environment, a high-dimensional ant robot must navigate from the bottom left to the top left of a maze, passing through hallways. 
The task is challenging due to its long duration, with episodes lasting 500 timesteps, and the considerable distance to be traversed. 
Compared to evaluation on goals both in top left room and in the central hallway in PEG, our evaluation only focuses on the ant reaching the most difficult four goals in the top left room.
Besides, we use all 32 goals of different positions in the maze to be the training goals returned by environment for \toolegc{}.
The maze itself measures approximately 6 x 8 meters. Success is determined by ensuring the L2 distance between the ant's $xy$ position and the goal is less than 1.0 meter, roughly the width of a cell in the maze. 
The Ant Maze environment features the highest dimensional state and goal spaces, totaling 29 dimensions. These include the ant's position, joint angles, and joint velocities. Specifically, the first three dimensions represent the $xyz$ position, the next 12 dimensions correspond to the joint angles of the ant's four limbs, and the remaining 14 dimensions capture the velocities in the $xy$ plane and of each joint. The action space consists of 8 dimensions, controlling the hip and ankle actuators of the ant's limbs.

\subsection{Point Maze} The 2D point agent starts at the bottom left corner of a 10 x 10 maze and is tasked with reaching the top right corner within 50 timesteps. 
The state space and action space are both two-dimensional, corresponding to the agent's position and velocity on the plane. Success is determined if the L2 distance between the agent's position and the goal is less than 0.15. 
This environment is directly adapted from \cite{pitis2020maximum} without any modifications. In \toolegc{}, the training goals from environments is randomly chose from 11 goals in different positions of maze.

\subsection{Block and Pen Rotation} The hand must manipulate both a thin pen or a block to achieve target rotations. 
Manipulating the thin pen presents a greater challenge than manipulating the block due to the pen's tendency to slip, requiring more precise control.
We utilize variant versions of the gymnasium environments: "HandManipulatePenRotate-v1" and "HandManipulateBlockRotateXYZ-v1". 
These environments introduce randomized target rotations for all axes of the block and x, y axes of the pen in each episode.
The state space for both tasks consists of 61 dimensions, providing details on the robot's joint and object states, as well as goal information. 
The goal space remains consistent at 7 dimensions, indicating the target pose information.
During evaluation, the latest policy is evaluated across 50 episodes for each task, with each episode featuring a distinct random goal.
In \toolegc{}, the training goals from environments is also randomly generated. Pen Rotation is particularly challenging due to the pen's thin structure, which requires precise control to prevent it from dropping. 
We intended to convey that this is the most difficult benchmark (with 61 observation space dimensions and 20 action space dimensions) in our test suite.

\section{Baselines}
\label{supp:baselines}

In this section, we present the pseudocode for all baseline methods. Note that, except for L3P, each baseline employs a different strategy for sampling data in the real environment within this framework. 
Therefore, we first display the general training framework for MBRL and subsequently provide the pseudocode for each baseline's data sampling method in the real environment.

\begin{algorithm}[H]
\begin{algorithmic}[1]
 \State\textbf{Input:} Policy $\pi^G$, $\pi^E$, Environment Goal Distribution $G$, World Model $\hat{M}$, reward function $r^G$, $r^E$
    \State $\mathcal{D} \gets \{ \}$ Initialize buffer. 
    \For{Episode $i=1$ to $N_{\text{train}}$} 
        \State \color{blue}$\tau \gets$  Collect trajectories$( \ldots )$\color{black}
       \State $\mathcal{D} \gets \mathcal{D} \cup \tau$
       \State Update model $\hat{M}$ with $(s_t, a_t, s_{t+1}) \sim \mathcal{D}$
       \State Update $\pi^G$ in imagination with $\hat{M}$ to maximize $r^G$
       \State Update $\pi^E$ in imagination with $\hat{M}$ to maximize $r^E$
       \EndFor
\end{algorithmic}
\caption{General MBRL Training Framework}
\label{alg:mbrl_training}
\end{algorithm}

\subsection{Go-Explore}
\label{supp:go_explore}

To enhance the exploration area of the explorer, we adopt the Go-Explore strategy (\cite{ecoffet2019go}). This approach initially employs a goal-conditioned policy, $\pi^G$, to approach a specified goal $g$ as closely as possible, a process referred to as the Go-phase. Following this, the explorer, $\pi^E$, is used to further explore the environment starting from the terminal state of the Go-phase, known as the Explore-phase.

The effectiveness of the trajectories generated by the Go-Explore strategy heavily depends on the choice of the goal $g$ during the Go-phase. Thus, establishing an efficient goal selection mechanism for the Go-phase is crucial. If the chosen goal $g$ is too easy, the explorer will not sufficiently explore the environment. Conversely, if the goal $g$ is too difficult, the goal-reaching policy $\pi^G$ will be unable to approach it effectively. Therefore, the objective is to develop a goal selection mechanism that identifies a goal $g$ capable of guiding the agent to a region with high exploration potential during the Go-phase. Below, we provide the pseudocode for the Go-Explore strategy.

\begin{algorithm}[H]
\begin{algorithmic}[1]
    \Function{GO-EXPLORE($g, \pi^G, \pi^E$)}{}
    \State $s_0 \gets$ env.reset()
    \State $\tau \gets \{s_0\}$
    \For{Step $t=1\ to\ T_{\text{Go}}$}
        \State $s_t \gets$  env.step($\pi^G(s_{t-1}, g)$)
        \State $\tau \gets \tau \cup \{s_t\}$
        \If{agent reach $g$}
        \State break
        \EndIf
    \EndFor
    \State $t_e = t$
    \For{Step $t=t_e\ to\ t_e + T_{\text{Explore}}$}
        \State $s_t \gets$  env.step($\pi^E(s_{t-1})$)
        \State $\tau \gets \tau \cup \{s_t\}$
    \EndFor
    \\\Return $\tau$
\EndFunction
\end{algorithmic}
\caption{Go Explore Framework}
\label{alg:go-explore}
\end{algorithm}

\subsection{GC-Dreamer}
GC-Dreamer follows a goal-conditioned approach where trajectories are collected by goal-conditioned policy $\pi^G$, and the goals are returned from the training environment. 
\begin{algorithm}[H]
\begin{algorithmic}[1]
\Function{Collect trajectories($\ldots$)}{}
\State $g \gets$ Returned by environment 
\State $\tau \gets$ Sample a trajectories by $\pi^G$ using goal $g$
    \\\Return $\tau$
\EndFunction
\end{algorithmic}
\caption{GC-Dreamer Goal Sampling}
\label{alg:gc-dreamer}
\end{algorithm}

\subsection{PEG}

PEG adopts a strategy where trajectories are collected by optimizing a specific equation using the MPPI method. The optimized goal is then used to guide exploration through the GO-EXPLORE algorithm.

\begin{algorithm}[H]
\begin{algorithmic}[1]
\Function{Collect trajectories($\ldots$)}{}
\State $g \gets$ Optimize Equation~\ref{eq:exp_potential_2} with MPPI
\State $\tau \gets \textit{GO-EXPLORE}(g, \pi^G, \pi^E)$
    \\\Return $\tau$
\EndFunction
\end{algorithmic}
\caption{PEG Sampling}
\label{alg:peg}
\end{algorithm}

\subsection{PEG-G}

PEG-G combines the utilization of goals from the environment with those generated by optimizing the equation using MPPI. This approach alternates between the two strategies based on the episode index.

\begin{algorithm}[H]
\begin{algorithmic}[1]
\Function{Collect trajectories($\ldots$)}{}
\If{episode $i \% 2 = 0$}
    \State $g \gets$ Optimize Equation~\ref{eq:exp_potential_2} with MPPI
    \State $\tau \gets \textit{GO-EXPLORE}(g, \pi^G, \pi^E)$
\Else
    \State $g \gets$ Returned by environment 
    \State $\tau \gets$ Sample a trajectories by $\pi^G$ using goal $g$
    \EndIf
    \\\Return $\tau$
\EndFunction
\end{algorithmic}
\caption{PEG-G Sampling}
\label{alg:PEG-G}
\end{algorithm}

\subsection{MEGA}

For model-based MEGA, we directly utilize the implementation method described in the PEG paper. 
This involves transplanting MEGA's KDE model and using a goal-conditioned value function within the LEXA framework to filter goals based on reachability. 
The PEG paper has demonstrated that their implementation of MEGA achieves superior performance compared to the original MEGA baseline.

\begin{algorithm}[H]
\begin{algorithmic}[1]
\Function{Collect trajectories($\ldots$)}{}
\State $g \gets \min_{g \in \mathcal{D}} \widehat{p}(g)$
\State $\tau \gets \textit{GO-EXPLORE}(g, \pi^G, \pi^E)$
    \\\Return $\tau$
\EndFunction
\end{algorithmic}
\caption{MEGA Goal Sampling}
\label{alg:mega}
\end{algorithm}

\subsection{MEGA-G}

Similar to PEG-G, MEGA-G alternates between using goals from the environment and MEGA goal picking strategy.

\begin{algorithm}[H]
\begin{algorithmic}[1]
\Function{Collect trajectories($\ldots$)}{}
\If{episode $i \% 2 = 0$}
    \State $g \gets \min_{g \in \mathcal{D}} \widehat{p}(g)$
    \State $\tau \gets \textit{GO-EXPLORE}(g, \pi^G, \pi^E)$
\Else
    \State $g \gets$ Returned by environment 
    \State $\tau \gets$ Sample a trajectories by $\pi^G$ using goal $g$
    \EndIf
    \\\Return $\tau$
\EndFunction
\end{algorithmic}
\caption{MEGA-G Goal Sampling}
\label{alg:MEGA-G}
\end{algorithm}

\subsection{MEGA+PEG}

MEGA+PEG combines the strategies of MEGA and PEG. this baseline firstly employs MEGA to sample a batch of candidate goals, all of which have low density in the replay buffer. 
Subsequently, their exploration potential is evaluated using PEG, with the most valuable one selected as the exploration goal.

\begin{algorithm}[H]
\begin{algorithmic}[1]
\Function{Collect trajectories($\ldots$)}{}

\State $G \gets$ \text{Top-10 smallest} $\widehat{p}(g) \text{ for } g \in \mathcal{D}$
\State $g \gets$ Optimize Equation~\ref{eq:exp_potential_2} for $g \in G$
\State $\tau \gets \textit{GO-EXPLORE}(g, \pi^G, \pi^E)$
    \\\Return $\tau$
\EndFunction
\end{algorithmic}
\caption{MEGA+PEG Goal Sampling}
\label{alg:mega+peg}
\end{algorithm}

\subsection{\toolwithoutpeg{}}

\toolwithoutpeg{} utilizes a goal-picking strategy based on Gaussian Mixture Model (GMM) clustering to generate goals for exploration without employing PEG optimization.
The goal picked by this method is sampled at the edge of our latent space clusters, which is the main contribution of our paper.

\begin{algorithm}[H]
\begin{algorithmic}[1]
\Function{Collect trajectories($\ldots$)}{}

\State $D_{exp} \leftarrow\{\}$
\For{episode $i = 1$ to $N_{\tau}$}

    \State $G_{candidate} \leftarrow$ Sample $N_{candidate}$ points from $GMM$
    \State $G_{edge} \leftarrow$ $N_{edge}$ points in $G_{candidate}$ with the smallest total probability of the $GMM$.
    \State $g^E \leftarrow$ Randomly select a $g$ from $G_{edge}$
    \State $\tau \leftarrow \textit{GO-EXPLORE}(g^E, \pi^G, \pi^E)$
        \\\Return $\tau$
    \EndFor
    
\EndFunction
\end{algorithmic}
\caption{\toolwithoutpeg{} Goal Sampling}
\label{alg:CE2-without-peg}
\end{algorithm}

\subsection{L3P}

Our implementation of L3P follows the original code provided in the L3P paper(\cite{zhang2021world}). For more details on the pseudocode and specific implementation, please refer to the descriptions in their paper.

\section{Implementation Details}
\label{supp:implementation_details}

\subsection{Farthest Point Sampling (FPS) Algorithm}
\label{subs: FPS}

\begin{algorithm}[H]
\caption{Farthest Point Sampling (FPS)}
\label{alg:FPS}
\begin{algorithmic}[1]
\Function{FPS}{points, num\_samples}
    \State sampled\_points $\leftarrow$ [ ]
    \State first\_point $\leftarrow$ random.choice(points)
    \State sampled\_points.append(first\_point)
    \State min\_distances $\leftarrow$ [float('inf')] $\times$ len(points)
    \For{each point $p$ in points}
        \State min\_distances[p] $\leftarrow$ distance(p, first\_point)
    \EndFor
    \For{iteration $i = 1$ to num\_samples-1}
        \State farthest\_point\_index $\leftarrow$ argmax(min\_distances)
        \State farthest\_point $\leftarrow$ points[farthest\_point\_index]
        \State sampled\_points.append(farthest\_point)
        \For{each point $p$ in points}
            \State min\_distances[p] $\leftarrow$ min(min\_distances[p], distance(p, farthest\_point))
        \EndFor
    \EndFor
    \State \Return sampled\_points
\EndFunction
\end{algorithmic}
\end{algorithm}

The Farthest Point Selection (FPS) algorithm, commonly employed in various applications including point cloud simplification and image sampling, initializes by creating an empty list termed 'sampled\_points' to retain the selected points. 
The process initiates by randomly selecting an initial point from the input point set, designated as 'points', and appending it to 'sampled\_points'. 
Subsequently, 'min\_distances' is initialized to track the minimum distance from each point to any of the sampled points, with initial values set to infinity.
The core procedure entails iteratively selecting points until reaching the desired number of samples. At each iteration, the algorithm identifies the point in 'points' with the maximum minimum distance to the previously sampled points and includes it in 'sampled\_points'. Concurrently, 'min\_distances' is updated to reflect the recalculated minimum distance of each point to any of the sampled points.
The algorithm incorporates two auxiliary functions: 'distance(point1, point2)', facilitating the computation of the Euclidean distance between two points, and 'argmax(array)', which returns the index of the maximum value within an array.
See pseudocode Algorithm~\ref{alg:FPS} for more details about FPS.

\subsection{Runtime}

\begin{table}[H]
\centering
\caption{Runtimes per experiment.}
\vspace{1ex}
\label{table:runtime}
\begin{tabular}{lcccc}
\toprule
& Total Runtime (Hours) & Total Steps & Episode Length & Seconds per Episode \\ 
\midrule
3-Block Stacking & 60 & 1e6 & 150 & 31.34 \\ 
Walker & 36 & 1e6 & 150 & 18.62 \\ 
Ant Maze & 56 & 1e6 & 500 & 96.91 \\ 
Point Maze & 36 & 1e6 & 50 & 5.60 \\ 
Block Rotation & 58 & 1e6 & 150 & 30.74 \\ 
Pen Rotation & 58 & 1e6 & 150 & 30.42 \\ 
\bottomrule
\end{tabular}
\end{table}

We conduct each experiment on GPU Nvidia A100 and require about 5GB of GPU memory. See Table~\ref{table:runtime} for specific running time of \tool{} for different task.
The running time of \toolegc{} has no big difference with \tool{}. The neural network updates of the policies and world model take most of runtime. However, \tool{} takes more time in goal selection compared to PEG and MEGA.
This is because \tool{} need evaluate the exploration potential of candidate goals sampled at the edge of latent clusters every time it picks a goal for exploration. 
In the PEG algorithm, the MPPI parameters are only updated at fixed intervals of multiple episodes. This means that the exploration potential is assessed for a batch of data after a certain number of episodes have been completed. 
In contrast, our method evaluates the exploration potential for a batch of candidate goals at each episode when sampling exploration goals.
However, we can also follow the PEG by evaluating the exploration potential of a batch of candidate goals after multiple episodes. 
Between these updates, we can select exploration goals from the previously evaluated set of candidate goals.
While this may sacrifice some algorithm performance, it can significantly reduce the time \tool{} requires to select goals. We tried this setting and compared the computation time needed to optimize goal states for launching the Go-Explore procedure among our \tool{}
 and the baseline methods MEGA and PEG in the 3-Block Stacking environment. The average wall clock time are recorded in the Table~\ref{table:runtime2}.

\begin{table}[H]
\centering
\caption{Computation time needed to optimize goal states}
\vspace{1ex}
\label{table:runtime2}
\begin{tabular}{lcccc}
\toprule
Method & Seconds/Episode \\ 
\midrule
\tool{} & 0.56 \\ 
PEG & 0.53 \\ 
MEGA & 0.47 \\ 
\bottomrule
\end{tabular}
\end{table}

\subsection{Hyperparameters}
\label{subsec: hyperparameters}
Similar to PEG, we use the default hyperparameters of the LEXA backbone MBRL agent (e.g., learning rate, optimizer, network architecture) and keep them consistent across all baselines.
For the Gaussian Mixture Model(GMM), we set the learning rate to be 3e-4, optimizer to be Adam. We also test result of different cluster number(10, 30, 50) setting as we show in ablation experiment.
Every time we sample points in the GMM, we set the point number $N_{candidate} = 1000$ and we set the edge point number $N_{edge} = 100$. 
After then, we evaluate the exploration potential of these 100 points and pick the point as goal state which have largest exploration potential value.
In addition, to improve clustering in the latent space, we reduced the latent space dimension from 400(LEXA setting) to 50. 
We tested the training speed and results with both a latent space dimension of 400 and 50. 
We found that a latent space dimension of 50 allows for faster clustering and accelerates the training process. 
Meanwhile, although a latent space dimension of 400 reduces the training speed a little, it does not affect the final success rate.

\section{Additional Experiments}

\subsection{Space explored image for 3-Block Stacking}
\label{subs: 3-block space explored}

\begin{figure}[h] 
  \centering
  \includegraphics[width=1\textwidth]{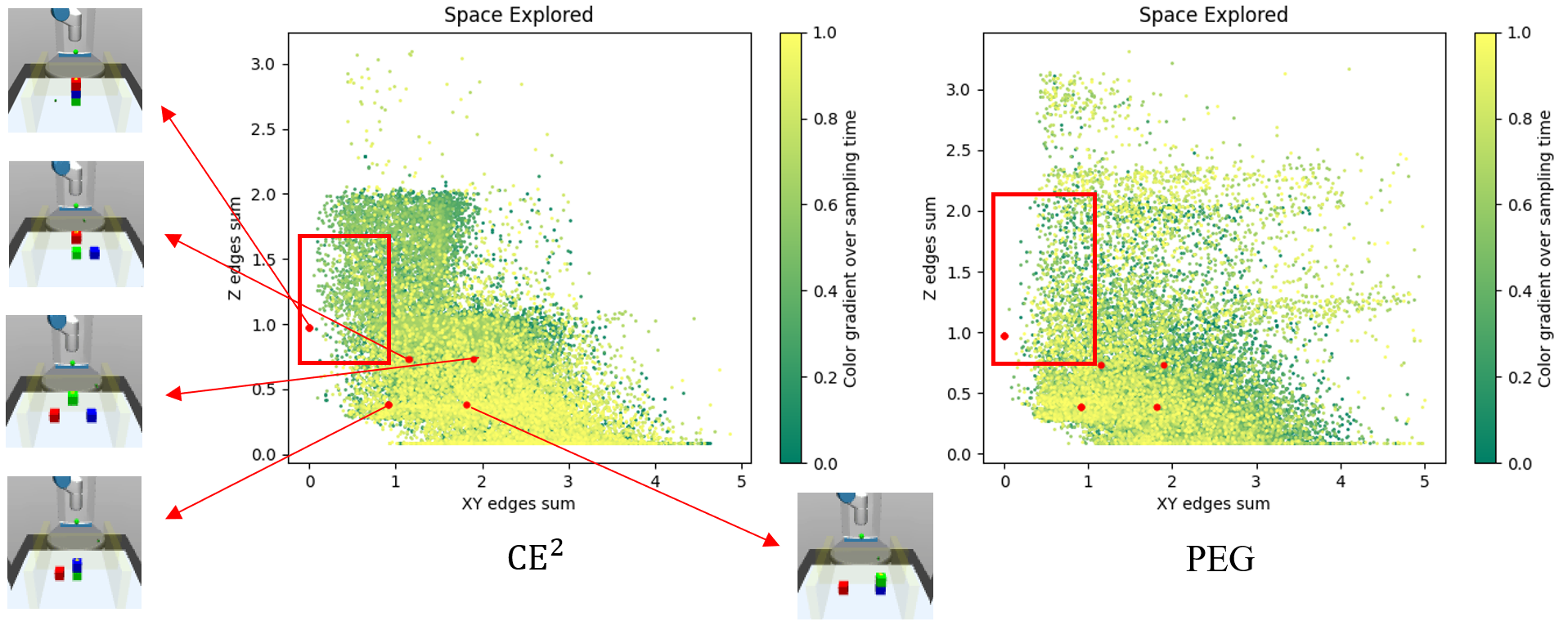}
  \caption{Space explored by \tool{} and PEG in the 3-Block Stacking environment at 1M steps. 
  X-axis: the sum of the three sides of the triangle projected on the x-y plane by the three block-connected triangles.
    Y-axis: sum of heights (z-coordinates) of the three blocks.
    Red points: evaluation goals.
    Other points: observations of trajectories sampled in real environment. Color from green to yellow means to be sampled more recent.}
  \label{fig:3-block space exlored}
\end{figure}

In 3-Block Stacking, we innovatively designed a method based on the coordinates of three blocks to demonstrate the degree of exploration in the environment, showed in Fig~\ref{fig:3-block space exlored}.
Firstly, we establish connections between the coordinates of three blocks in three-dimensional space, forming a spatial triangle.
This spatial triangle serves to express the relative positions and distances of the three blocks within the space.
Subsequently, we project this spatial triangle onto the xy-plane. The summation of the lengths of the sides of the projected triangle on the xy-plane reflects the dispersal of the three blocks within the xy-plane, 
while the total sum of the z-coordinates of the three blocks indicates their relative positions in height
Utilizing the former as the x-axis and the latter as the y-axis, we depict a schematic illustration of the spatial exploration of 3-Block Stacking. 
We observe that \tool{} exhibits a more targeted and in-depth exploration around the target space (highlighted within the red box) compared to PEG.
Simultaneously, we observed that PEG tends to conduct numerous explorations in the upper-left region of the exploration graph, 
which are often futile and irrelevant to the goal of completing block stacking. 
This highlights the advantage of \tool{}, which benefits from the constraints imposed by clustering and avoids blindly exploring areas. 
Moreover, sampling at the edges of clusters ensures the profitability of exploration, enabling more efficient exploration of the vicinity of the goal space compared to PEG.

\subsection{More Exploration Process}
\label{subs: MoreAntExploreSpace}

\begin{figure}[h]
  \centering
  \includegraphics[width=0.9\textwidth]{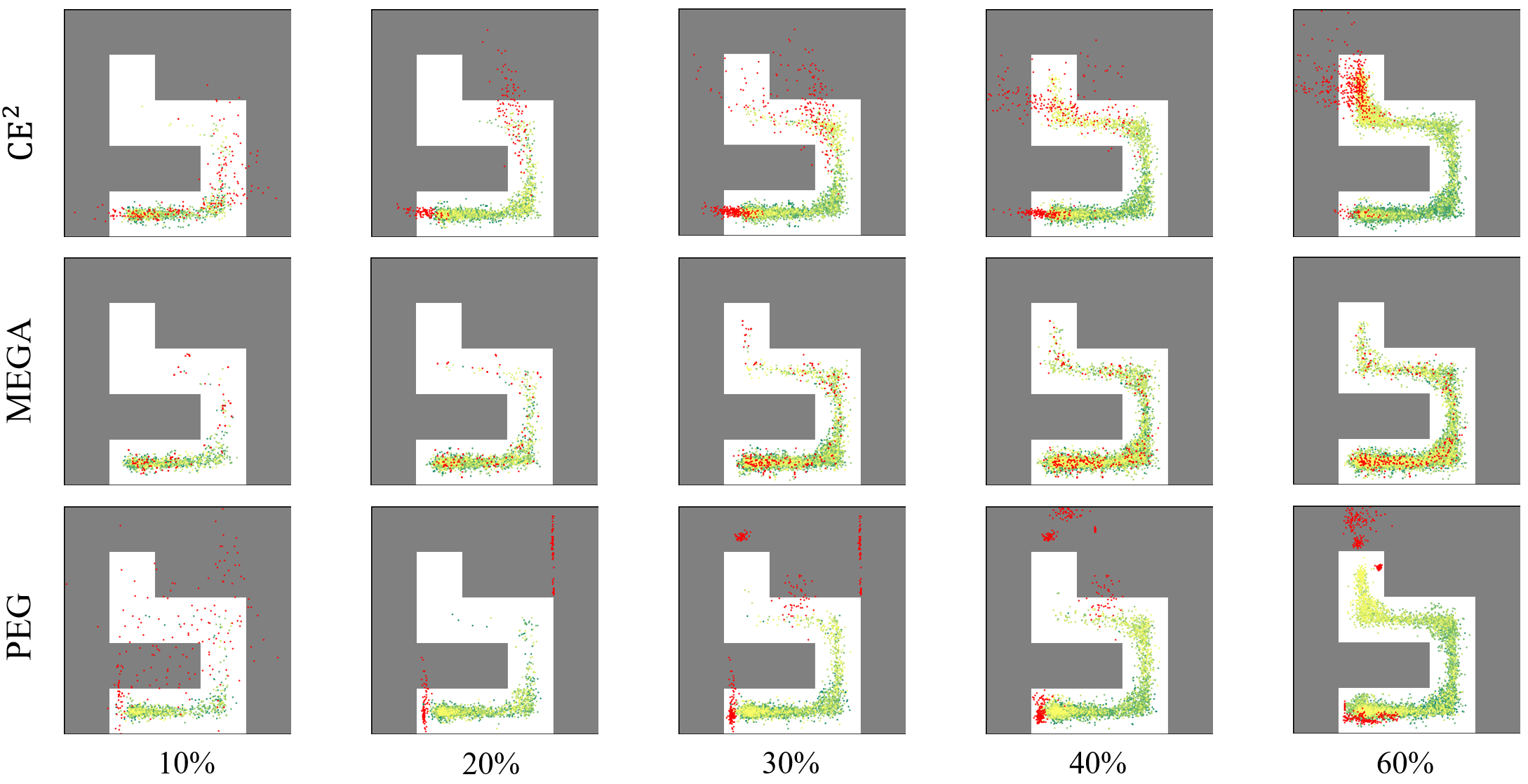}
  \caption{Comparison of exploration goals generated by \tool{}, MEGA and PEG}
  \label{fig:MoreAntExploreSpace}
\end{figure}

We present a comparison of exploration targets generated by \tool{}, MEGA and PEG approaches over the training process in the Ant Maze environment.
In the Fig~\ref{fig:MoreAntExploreSpace}, red points represent the generated exploration targets by different methods. 
We observe that the exploration targets generated by \tool{} are significantly superior to those generated by MEGA and PEG. 
Specifically, \tool{} consistently generates points located at the forefront of agent exploration and within the agent's reachable capability range. 
In contrast, the targets generated by MEGA exhibit greater dispersion and sparsity, which are disadvantageous for concentrated exploration of forefront regions. 
Moreover, PEG consistently generates targets outside the Maze channels, rendering these exploration targets not only far beyond the agent's capability range but also meaningless.

\subsection{Centroids Visualization}
\label{susb: CentroidsVisualization}

\begin{figure}[h]
  \centering
  \includegraphics[width=0.9\textwidth]{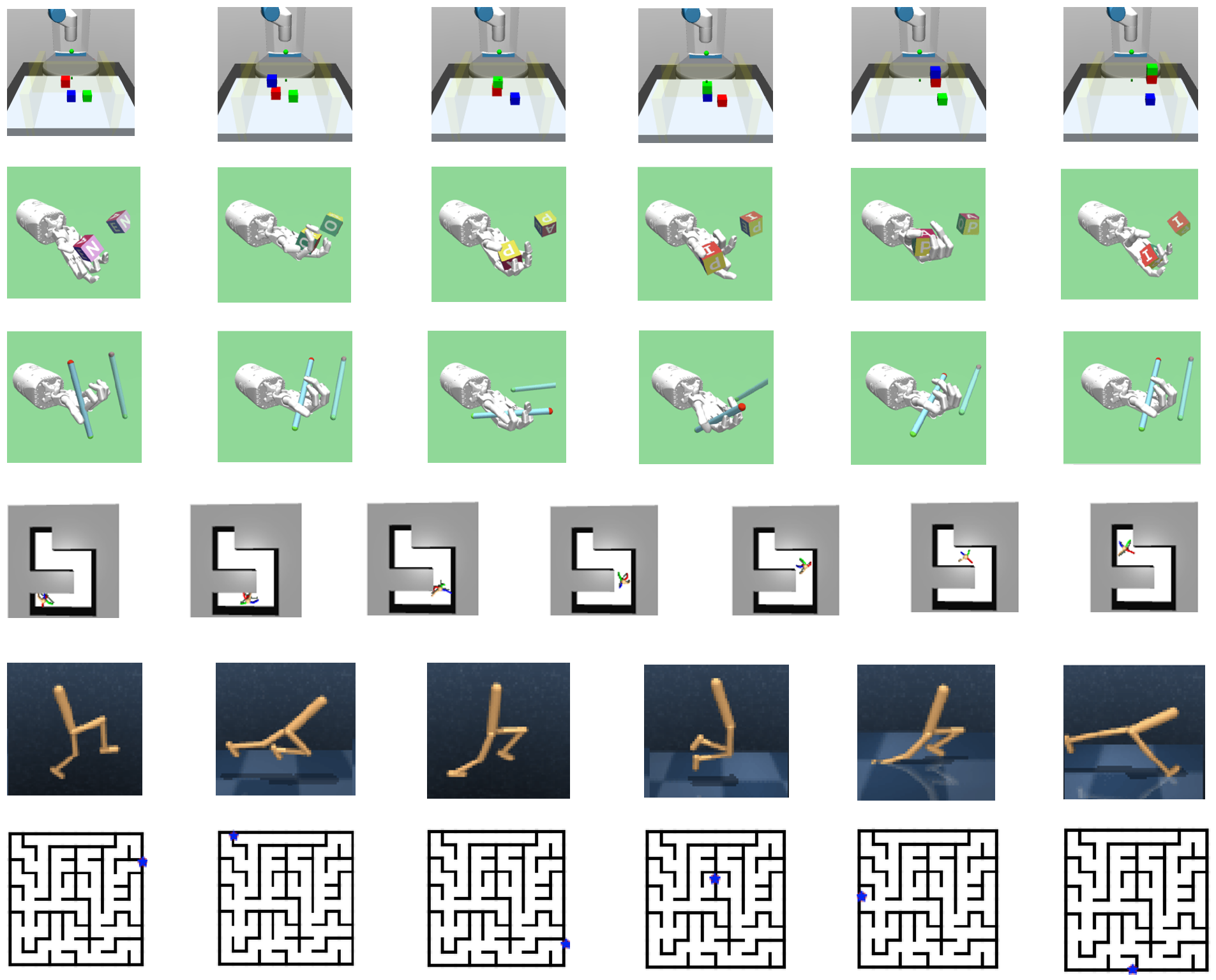}
  \caption{some centroids visualization of GMMs in \tool{}.}
  \label{fig:moresubgoals}
\end{figure}

By decoding the centroids of GMMs in latent space, we can visualize some centroids of GMMs in \tool{}, showing in Fig.~\ref{fig:moresubgoals}

\subsection{Full results for \toolegc{}}
\label{subs: fullresultforegc}

\begin{figure}[t] 
  \centering
  \subfigure[Point Maze]{\includegraphics[width=0.3\textwidth]{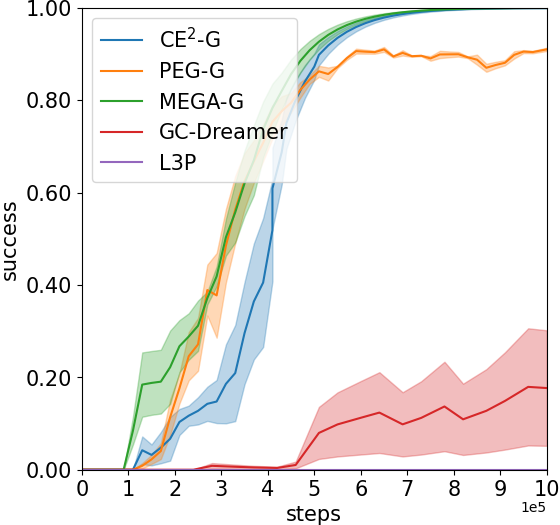}}
  \subfigure[Ant Maze]{\includegraphics[width=0.3\textwidth]{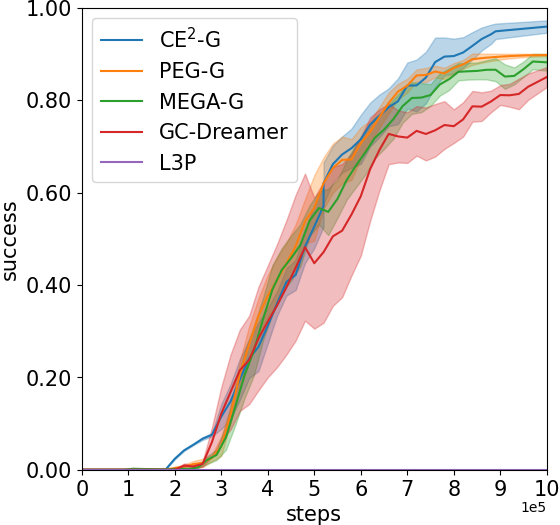}}
    \subfigure[Walker]{\includegraphics[width=0.3\textwidth]{Exp_Imgs/Walker-egc.png}}
  \subfigure[3-Block Stacking]{\includegraphics[width=0.3\textwidth]{Exp_Imgs/3-Block-hardgoal-egc.png}}
  \subfigure[Block Rotation]{\includegraphics[width=0.3\textwidth]{Exp_Imgs/Rotate_Block-egc.png}}
  \subfigure[Pen Rotation]{\includegraphics[width=0.3\textwidth]{Exp_Imgs/Rotate_Pen-egc.png}}
  \caption{Full experiment Results comparing \toolegc{} with the baselines in six environments.}
  \label{fig:fullresultforegc}
\end{figure}

Please see Fig.~\ref{fig:fullresultforegc}

\subsection{More Ablation Experiments}
\label{subs: moreablation}

\begin{figure}[t] 
  \centering
  \subfigure[3 Block]{\includegraphics[width=0.32\textwidth]{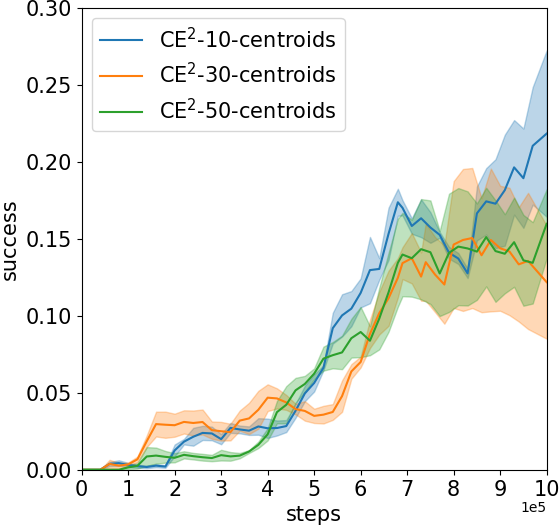}}
  \subfigure[Ant Maze]{\includegraphics[width=0.32\textwidth]{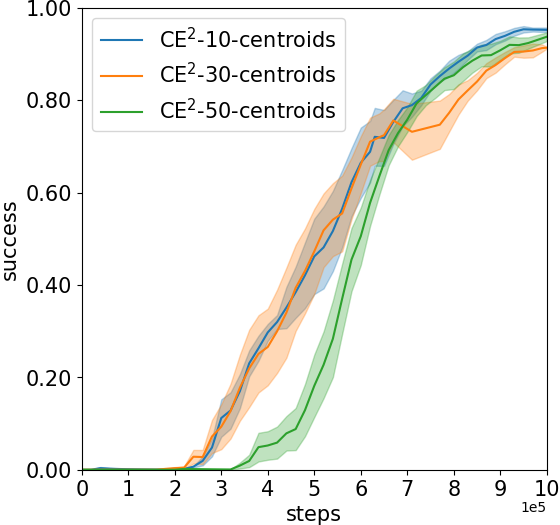}}
  \subfigure[Point Maze]{\includegraphics[width=0.32\textwidth]{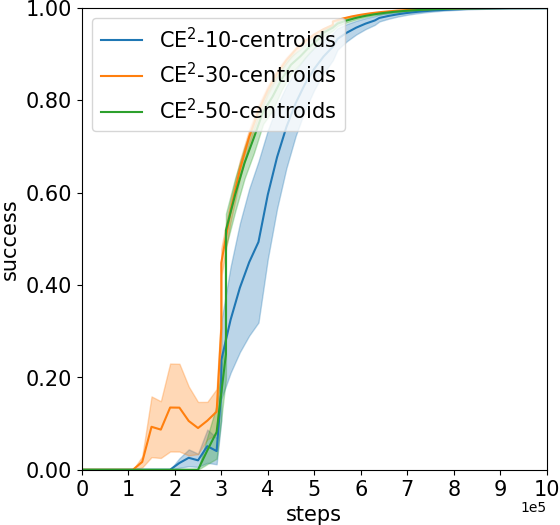}}
  \subfigure[Block Rotation]{\includegraphics[width=0.32\textwidth]{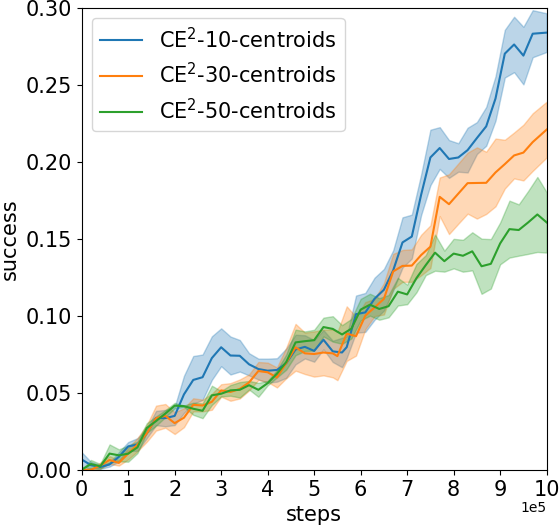}}
  \subfigure[Pen Rotation]{\includegraphics[width=0.32\textwidth]{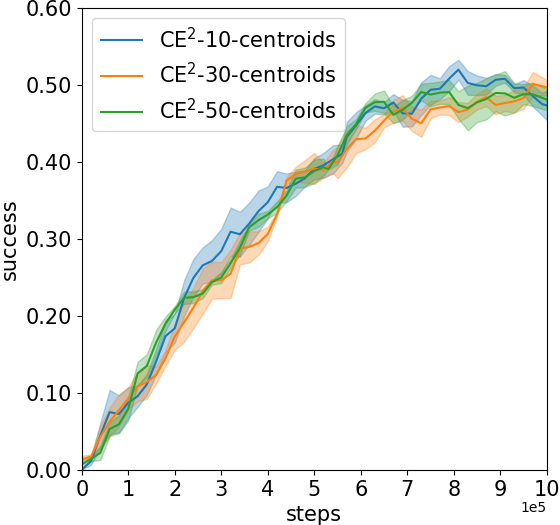}}
  \subfigure[Walker]{\includegraphics[width=0.32\textwidth]{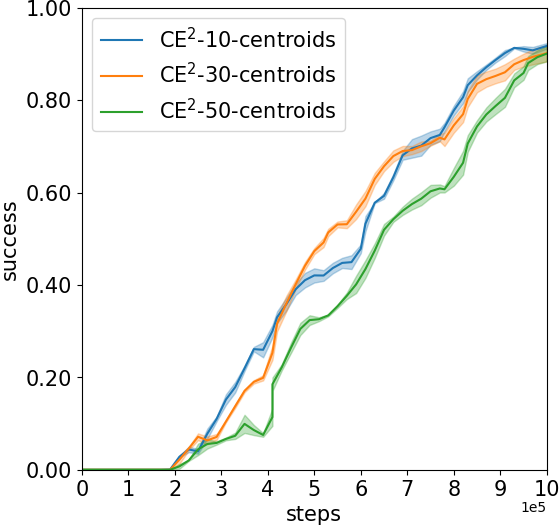}}
  
  \caption{Ablation Study with Different Cluster Number.}
  \label{fig:exp_abl_2}
\end{figure}

We conducted experiments in the \tool{} with different numbers (10, 30, 50) of clusters to observe if the results are sensitive to the number of clusters. 
The results are shown in Fig~\ref{fig:exp_abl_2}. We found that the performance of \tool{} is not strongly correlated with the number of clusters; 
as long as the number of clusters is sufficient to represent key state regions, the results tend to be stable, which demonstrates the robustness of our approach.

%%%%%%%%%%%%%%%%%%%%%%%%%%%%%%%%%%%%%%%%%%%%%%%%%%%%%%%%%%%%
\clearpage
\newpage
\section*{NeurIPS Paper Checklist}

\begin{enumerate}

\item {\bf Claims}
    \item[] Question: Do the main claims made in the abstract and introduction accurately reflect the paper's contributions and scope?
    \item[] Answer: \answerYes{} % Replace by \answerYes{}, \answerNo{}, or \answerNA{}.
    \item[] Justification: Our abstract and introduction clearly articulate the primary claims of the paper, specifically focusing on the development and implementation of the \tool{} algorithm for goal-directed exploration in goal-conditioned reinforcement learning (GCRL). The introduction outlines the main contributions, including the proposal of a new exploration mechanism, the clustering strategy to prioritize accessible goals, and the validation of \tool{} in various challenging robotics environments. These claims are aligned with the theoretical framework and experimental results presented in the paper, ensuring they accurately reflect the contributions and scope of the paper.

\item {\bf Limitations}
    \item[] Question: Does the paper discuss the limitations of the work performed by the authors?
    \item[] Answer: \answerYes{} % Replace by \answerYes{}, \answerNo{}, or \answerNA{}.
    \item[] Justification: Our paper clearly outlines several limitations of the proposed \tool{} algorithm in Appendix. 
    Firstly, it emphasizes the dependency on a well-trained latent space, which must accurately reflect reachable distances between states and facilitate dynamic prediction. 
    Training this latent space requires a robust temporal distance predictor, which we address by using the predictor network from LEXA, trained with simulated trajectories for stability. 
    Additionally, we discuss that \tool{} may face challenges with more complex tasks such as peg insertion or fluid tasks in ManiSkill2. 

\item {\bf Theory Assumptions and Proofs}
    \item[] Question: For each theoretical result, does the paper provide the full set of assumptions and a complete (and correct) proof?
    \item[] Answer: \answerNA{} % Replace by \answerYes{}, \answerNo{}, or \answerNA{}.
    \item[] Justification: We does not include theoretical results

    \item {\bf Experimental Result Reproducibility}
    \item[] Question: Does the paper fully disclose all the information needed to reproduce the main experimental results of the paper to the extent that it affects the main claims and/or conclusions of the paper (regardless of whether the code and data are provided or not)?
    \item[] Answer: \answerYes{} % Replace by \answerYes{}, \answerNo{}, or \answerNA{}.
    \item[] Justification: In the Experiment section and appendix of our paper, we provide detailed descriptions of our experimental procedures and configurations. 
    This includes elucidating the origins and modifications made to all testing environments. We also present the pseudocode and implementation methods for all baseline models. 
    Additionally, we specify the devices and memory resources utilized, as well as enumerate the exact numerical values of the hyperparameters employed. 
    Moreover, we have made our code openly accessible. For further details, please refer to the Reproducibility Statement section.

\item {\bf Open access to data and code}
    \item[] Question: Does the paper provide open access to the data and code, with sufficient instructions to faithfully reproduce the main experimental results, as described in supplemental material?
    \item[] Answer: \answerYes{} % Replace by \answerYes{}, \answerNo{}, or \answerNA{}.
    \item[] Justification: As we answer in the previous question. We have open source our code and provide detailed instructions to reproduce the main experimental results. We illustrate the benchmark source, baseline settings and \tool{} implementation details. 

\item {\bf Experimental Setting/Details}
    \item[] Question: Does the paper specify all the training and test details (e.g., data splits, hyperparameters, how they were chosen, type of optimizer, etc.) necessary to understand the results?
    \item[] Answer: \answerYes{} % Replace by \answerYes{}, \answerNo{}, or \answerNA{}.
    \item[] Justification: We provide detailed descriptions of the hyperparameters used for \tool{} in Appendix, such as varying cluster numbers and latent space dimensions. 
    We also desbribe the setting of training, including learning rates, optimizers, and network architectures. This information is crucial for understanding and reproducing our experimental results. 

\item {\bf Experiment Statistical Significance}
    \item[] Question: Does the paper report error bars suitably and correctly defined or other appropriate information about the statistical significance of the experiments?
    \item[] Answer: \answerYes{} % Replace by \answerYes{}, \answerNo{}, or \answerNA{}.
    \item[] Justification: We repeated each experiment at least five times using different random seeds, and when plotting the results.
    As we showing in the Experiment section, we displayed the experimental error. 
    The solid line represents the average success rate, while the shaded region represents the standard deviation between the repeated experimental results.

\item {\bf Experiments Compute Resources}
    \item[] Question: For each experiment, does the paper provide sufficient information on the computer resources (type of compute workers, memory, time of execution) needed to reproduce the experiments?
    \item[] Answer: \answerYes{} % Replace by \answerYes{}, \answerNo{}, or \answerNA{}.
    \item[] Justification:  We clearly specifies the computer resources(Nvidia A100 GPU) and the amount of GPU memory required (approximately 5GB) in Appendix. 
    Additionally, we provides detailed information on the runtime of each experiment, including specific time metrics such as episode length and seconds per episode.
    
    
\item {\bf Code Of Ethics}
    \item[] Question: Does the research conducted in the paper conform, in every respect, with the NeurIPS Code of Ethics \url{https://neurips.cc/public/EthicsGuidelines}?
    \item[] Answer: \answerYes{} % Replace by \answerYes{}, \answerNo{}, or \answerNA{}.
    \item[] Justification: The research conducted in our paper aligns with the NeurIPS Code of Ethics. 
    We have reviewed the guidelines and ensured that our research adheres to ethical standards. 
    Additionally, we have taken measures to preserve anonymity and comply with relevant laws and regulations.

\item {\bf Broader Impacts}
    \item[] Question: Does the paper discuss both potential positive societal impacts and negative societal impacts of the work performed?
    \item[] Answer: \answerNA{} % Replace by \answerYes{}, \answerNo{}, or \answerNA{}.
    \item[] Justification: Our study focuses on solving exploration issues in the GCRL environment. At this stage, it remains largely theoretical and has negligible societal implications.
    
\item {\bf Safeguards}
    \item[] Question: Does the paper describe safeguards that have been put in place for responsible release of data or models that have a high risk for misuse (e.g., pretrained language models, image generators, or scraped datasets)?
    \item[] Answer: \answerNA{} % Replace by \answerYes{}, \answerNo{}, or \answerNA{}.
    \item[] Justification: Our paper poses no such risks.

\item {\bf Licenses for existing assets}
    \item[] Question: Are the creators or original owners of assets (e.g., code, data, models), used in the paper, properly credited and are the license and terms of use explicitly mentioned and properly respected?
    \item[] Answer: \answerYes{} % Replace by \answerYes{}, \answerNo{}, or \answerNA{}.
    \item[] Justification: Our paper properly credits the creators or original owners of assets used, including code, data, and models. 
    The licenses and terms of use are explicitly respected. Specifically, we cite the original papers for code packages or datasets used, state the version of the assets, and include URLs where possible.

\item {\bf New Assets}
    \item[] Question: Are new assets introduced in the paper well documented and is the documentation provided alongside the assets?
    \item[] Answer: \answerYes{} % Replace by \answerYes{}, \answerNo{}, or \answerNA{}.
    \item[] Justification: We have documented our code and provided detailed instructions on its usage, licenses, and permissible scope of use. 
    Additionally, we have included the documentation alongside the assets to ensure accessibility and clarity for users.

\item {\bf Crowdsourcing and Research with Human Subjects}
    \item[] Question: For crowdsourcing experiments and research with human subjects, does the paper include the full text of instructions given to participants and screenshots, if applicable, as well as details about compensation (if any)? 
    \item[] Answer: \answerNA{} % Replace by \answerYes{}, \answerNo{}, or \answerNA{}.
    \item[] Justification: Our paper not involve crowdsourcing nor research with human subjects.

\item {\bf Institutional Review Board (IRB) Approvals or Equivalent for Research with Human Subjects}
    \item[] Question: Does the paper describe potential risks incurred by study participants, whether such risks were disclosed to the subjects, and whether Institutional Review Board (IRB) approvals (or an equivalent approval/review based on the requirements of your country or institution) were obtained?
    \item[] Answer: \answerNA{} % Replace by \answerYes{}, \answerNo{}, or \answerNA{}.
    \item[] Justification: Our paper does not involve crowdsourcing nor research with human subjects.

\end{enumerate}
\end{document}